\documentclass[review]{elsarticle}
\usepackage{hyperref}
\usepackage{float}
\usepackage{hhline}
\usepackage{multirow}
\usepackage{subfig}
\usepackage{textcomp}
\usepackage{color}
\usepackage{amsmath}
\usepackage{graphicx}
\usepackage{csquotes}
\usepackage{xcolor}

\begin{document}

\begin{frontmatter}

\title{Damage detection in operational wind turbine blades using a new approach based on machine learning}

\author[add1]{Kartik Chandrasekhar\corref{cor1}}
\ead{k.chandrasekhar@sheffield.ac.uk}
\author[add2]{Nevena Stevanovic\address{add2}}
\author[add1]{Elizabeth J. Cross\address{add1}}
\author[add1]{Nikolaos Dervilis\address{add1}}
\author[add1]{Keith Worden\address{add1}}

\cortext[cor1]{Corresponding author}
\address[add1]{Dynamics Research Group, Department of Mechanical Engineering, University of Sheffield, Mappin Street, Sheffield, S1 3JD, UK.}
\address[add2]{Siemens Gamesa Renewable Energy A/S, Borupvej 16, 7330 Brande, Denmark}

\begin{abstract}
The application of reliable structural health monitoring (SHM) technologies to operational wind turbine blades is a challenging task, due to the uncertain nature of the environments they operate in. In this paper, a novel SHM methodology, which uses Gaussian Processes (GPs) is proposed. The methodology takes advantage of the fact that the blades on a turbine are nominally identical in structural properties and encounter the same environmental and operational variables (EOVs). The properties of interest are the first edgewise frequencies of the blades. The GPs are used to predict the edge frequencies of one blade given that of another, after these relationships between the pairs of blades have been learned when the blades are in a healthy state. In using this approach, the proposed SHM methodology is able to identify when the blades start behaving differently from one another over time. To validate the concept, the proposed SHM system is applied to real onshore wind turbine blade data, where some form of damage was known to have taken place. X-bar control chart analysis of the residual errors between the GP predictions and actual frequencies show that the system successfully identified early onset of damage as early as six months before it was identified and remedied.
\end{abstract}


\end{frontmatter}

\section{Introduction} \label{sec:introduction}

Wind turbines rely on a number of structurally-critical components that may directly affect their power generation capabilities. Of note are the tower, the gearbox and generator (for variable-speed wind turbines), the magnetic drive (for direct-drive wind turbines), various bearings (such as main bearings), and perhaps most importantly, the blades \cite{marquez2012condition}. Blades are among the most expensive components - depending on their size; their manufacturing costs range between 10\% and 20\% of total manufacturing costs \cite{li2015review}. In recent years, blades have progressively become larger, to harvest more energy, and maximise power production. At the same time, since larger blades are associated with larger weight penalties, blade designs have continually been modified to keep the penalties low. These compromises have led to more flexible blades, and therefore, lower safety margins \cite{ciang2008structural, yang2013testing}.

Cracks and delaminations are reported to be the most common form of damage in blades \cite{ciang2008structural}; these typically occur close to the blade root since this is the region where the blades experience the highest bending strains due to turbine rotation; this is especially true when taking large blades into consideration. Towards the end of the design lifetime of the blades, damage mechanisms become accentuated during excessive levels of winds, lightning strikes, and ice accumulation (which leads to rotor imbalance problems). It is extremely important to avoid critical blade failures, because when damaged blades liberate, they have the potential to damage not only the turbines they were attached to, but also other turbines in their vicinity \cite{raivsutis2008review}.

There have also been heavy investments in the offshore wind industry, whereby wind turbines have increasingly been deployed on various seas and oceans, where wind conditions are favourable for energy production \cite{kaldellis2013shifting}. However, this freedom has come at a cost of ease of maintenance, since it is significantly more expensive to send technicians to inspect offshore structures via expensive transportation \cite{rockmann2017operation}. 

For all these reasons, Structural Health Monitoring (SHM) technologies have become appealing to both wind turbine manufacturers and operators alike. The use of SHM technologies has largely been driven by a number of factors: 

\begin{enumerate}
  \item minimisation of downtime,
  \item maximisation of reliability,
 \item  minimisation of operation and maintenance costs, and 
  \item extension of design lifetimes of monitored structures.
\end{enumerate}

Currently, most of the focus in SHM research on small-to-medium sized blades has been in laboratory environments, where conditions are controlled. Additionally, researchers have had the luxury of multiple sensors and data acquisition systems with high sampling rates. Signals captured by these sensors include strains \cite{rumsey2008structural, sierra2016damage}, velocities \cite{ghoshal2000structural}, accelerations \cite{adams2011structural}, acoustic emissions \cite{rumsey2008structural, lee2011US1, tang2016US2}, guided waves \cite{dervilis2014damage, taylor2014incipient}, thermographic images \cite{galleguillos2015thermographic}, and digital image correlation (DIC) images \cite{niezrecki2014inspection}, to name a few. This type of monitoring has facilitated high-fidelity diagnostics (indication and location of damage), as well as prognostics (remaining life prediction). There have been numerous methods and procedures to aid in damage indication and localisation; these include changes in stiffness \cite{rumsey2008experimental, paquette2008fatigue} and modal properties \cite{yang2015condition}, principal component analysis \cite{sierra2016damage}, image processing of thermographic \cite{galleguillos2015thermographic} and DIC images \cite{niezrecki2014inspection}, etc. Machine learning algorithms, such as artificial neural networks, have also been used previously \cite{dervilis2014damage}.

The above-mentioned research subjects are extremely valuable, as they illustrate various routes to tackle the ultimate goal of having a reliable SHM system. However, operational wind turbines around the world face challenges that cannot be replicated in laboratory environments. Due to this issue - at least for the time being - certain methods can seem to be impractical. Some of the challenges are listed below.

\begin{itemize}
\item Firstly, there are significant costs involved in applying numerous sensors with high data-acquisition rates; these include the costs of the sensors themselves, as well as costs associated with data storage and processing. One must also consider the fact that there are hundreds of thousands of wind turbines spread across the world.
\item Furthermore, monitoring and diagnostic systems need to be designed to be robust to the heavily nonstationary conditions that wind turbines face. Nonstationarity primarily stems from constantly-changing wind and loading conditions (gusts, turbulence, etc.), and taking these into account is therefore a challenging task.
\item Damage localisation is another challenge in operational wind turbines, since this is a function of excitation frequency and number of sensors, as well as data-acquisition rates and sensor placement. There are additional limitations as well; for example, one cannot place piezoelectric sensors along the length of the blades due to the risks associated with electromagnetic interference and lightning strikes.
\item Diagnostic methods also need to find a balance between true and false positives, and this would involve a cost-benefit analysis. For example, in a laboratory, a 3\% false positive rate may be deemed acceptable, but in the real world, if all turbines are stopped for 3\% of the time, costs associated with downtime will have to be borne by turbine operators. On the other hand, if it does turn out that a damage indicator is correct, and the damage becomes irreparable, costs associated with blade replacement can also be considerable.
\end{itemize}

This article proposes methodologies that address some of these challenges. As mentioned above, when turbines are operating, the effects of environmental and operational variables cause significant nonstationarity; this is especially true of the frequency response of the blades. For example, high wind speeds can cause faster rotation of the turbines, which lead to stiffer blades and higher natural frequencies; higher temperatures lead to less stiff blades, hence lowering the natural frequencies. The methodologies proposed in this article take advantage of the fact that the EOVs act simultaneously on all three blades. Therefore, when the monitored structural frequencies are viewed relative to each other, the complex nonstationarities are transformed into a simpler, and more explicable space. As a result of this transformation, it is possible to learn relationships between these monitored variables (as well as other EOVs). These relationships are learnt, in pairs, using probabilistic regression in the form of Gaussian Processes (GPs). The GPs are then used for predictive purposes, and the residuals between the actual signals and predicted signals can be used as an informative damage indicator.

The SHM scheme proposed is summarised in the next section. This is followed by a detailed review of the theory behind GPs. The article is complemented by three case studies of onshore wind turbines, where the proposed methodologies were applied, and would have successfully identified damage in advance of critical failure. It should be noted that the versatility of this methodology means that they can also be applied to offshore wind turbines.

\section{Methodology and Theory}\label{sec:methodology}

\begin{figure}[H]
	\vspace{0pt}
	\centering
	{\includegraphics[scale=0.5, trim=5cm 7cm 3.8cm 8cm, clip=true]{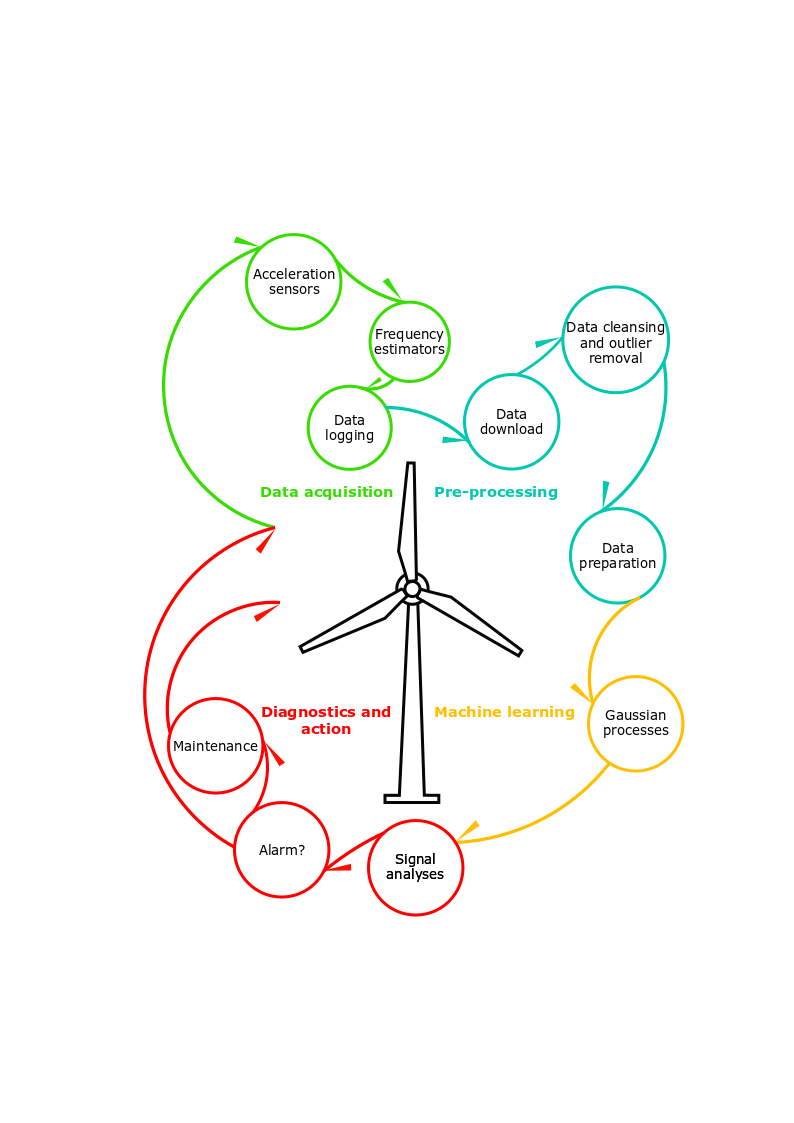}}
	\vspace{0in}
	\caption{Summary of proposed SHM system}
	\label{fig:summary_of_process}
\end{figure}

Figure \ref{fig:summary_of_process} illustrates a summary of the SHM system that is presented in this article. 

As shown in Figure \ref{fig:summary_of_process}, the SHM system capability is divided into four categories: data acquisition, pre-processing, machine learning, and diagnostics and actions; these are discussed in detail next.

\subsection{Data acquisition}

Acceleration sensors, and sensors that measure the environmental and operational variables (EOVs) measure the respective signals. The edge frequencies of each blade (i.e.\ the frequencies of the first bending mode in the blade edgewise direction) are estimated from the acceleration signals. In this paper, the machine learning techniques are applied to Supervisory Control and Data Acquisition (SCADA) data. For readers unfamiliar with the SCADA system, it is a utility that acquires data from sensors mounted on a structure, and sends them to a central computer for monitoring and control purposes. The data statistics are logged at regular intervals, and these are typically ten-minute averages \cite{wang2014scada}, which implies that the SCADA data has a low-pass filter applied to it. The variables used from the SCADA data include:

\begin{enumerate}
\item edge frequencies of each blade (in Hz),
\item ambient temperature (in \textdegree{C}),
\item power output (in W),
\item generator rotation speed (in RPM), and
\item pitch angle (in \textdegree).
\end{enumerate}

Prior to data logging into the SCADA system, the individual signals are usually sampled at a rate of 10 Hz (in some cases, higher sampling is also possible). The aim of this work is to detect long term damage, which typically occurs over the course of several weeks or months. Therefore, although the data in the SCADA system have a low-pass filter applied to them, they are still capable of identifying the long term degradation in the blade stiffness, and hence the edge frequency. The benefit of using SCADA data over the higher-resolution data is that it saves processing time quite considerably.

\subsection{Pre-processing}

In this part of the system, SCADA data from the respective channels were downloaded from a number of wind turbines whose blades were known to have incurred damage. These time-series were then filtered, and processed to remove outliers within the data. Outliers are quite a common occurrence in SCADA data, and typically occur in the data logging process. Furthermore, since the time-series are logged in various dedicated channels, occasionally, it is common for data to not be recorded at certain times in a given channel, whilst data in other channels are being logged. It is possible to resample the data such that all data channels have identical sampling rates. However, this requires some form of interpolation, which could result in misleading information. Therefore, it was decided that only data that had common timestamps would used for the remainder of the processing that followed.

Note that all the data used for the purposes of this article have been normalised prior to the application of the machine learning and signal processing methods. This is performed according to,

\begin{equation}\label{eq:data_norm}
x_{norm} = \frac{x - \mu}{\sigma}
\end{equation} 

\noindent where the $x$ are the data being normalised, $\mu$ is the mean, and $\sigma$ is the standard deviation of selected training data. The selection of training data will be discussed shortly, but it is the same training data that will be used in the training of the Gaussian Processes (GPs). There are two primary reasons for normalising the data.

\begin{itemize}
	\item To avoid numerical instabilities that may occur in various computations in GP inference.
	\item To avoid disclosing confidential information regarding blade properties. 
\end{itemize}

\begin{figure}[H]
  \vspace{0pt}
  \centering
   \subfloat[]{\includegraphics[scale=0.405, trim=0cm 0cm 0cm 0cm, clip=true]{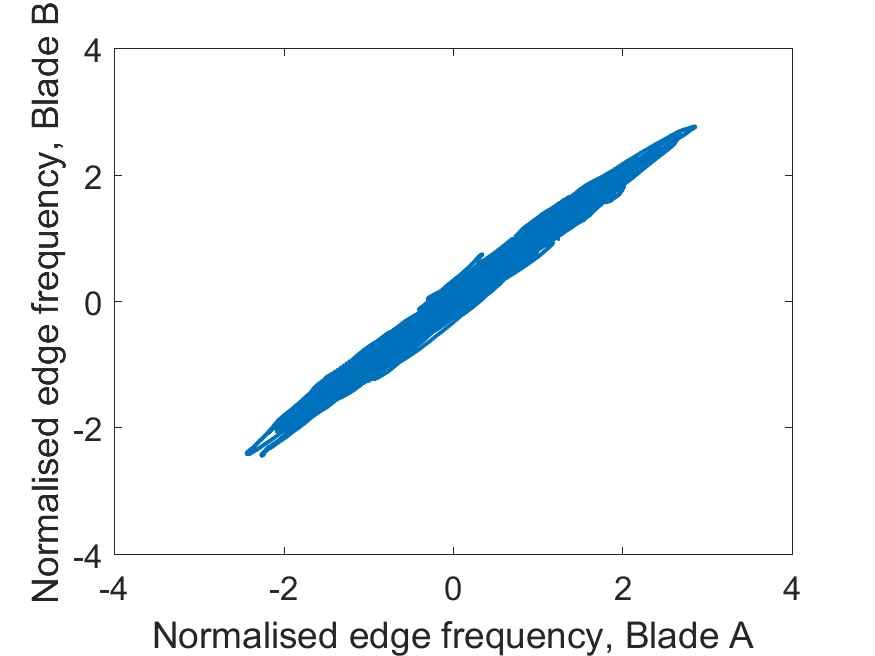}}
  \hspace{0in}
   \subfloat[]{\includegraphics[scale=0.405, trim=0cm 0cm 0cm 0cm, clip=true]{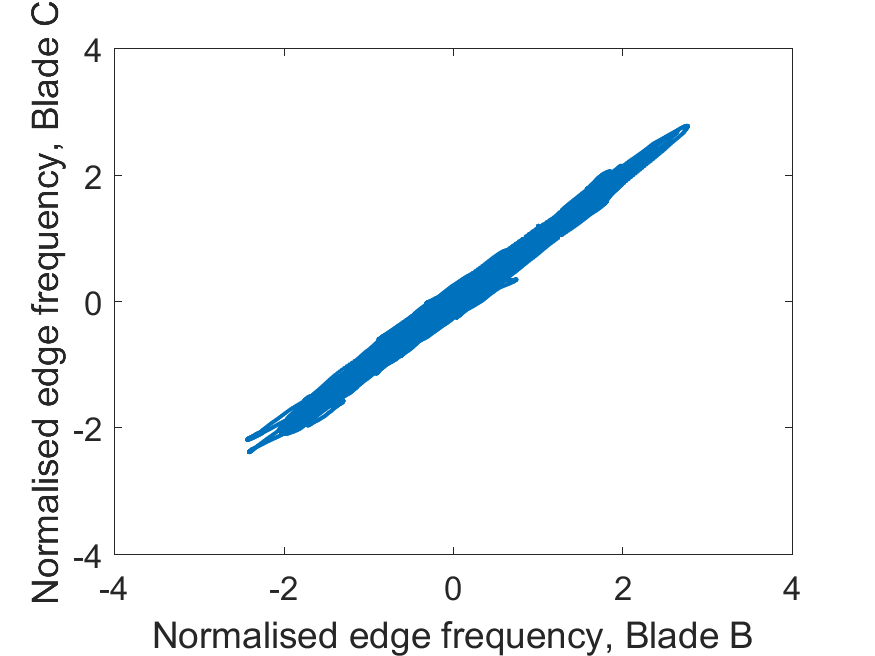}}
    \hspace{0in}
   \subfloat[]{\includegraphics[scale=0.405, trim=0cm 0cm 0cm 0cm, clip=true]{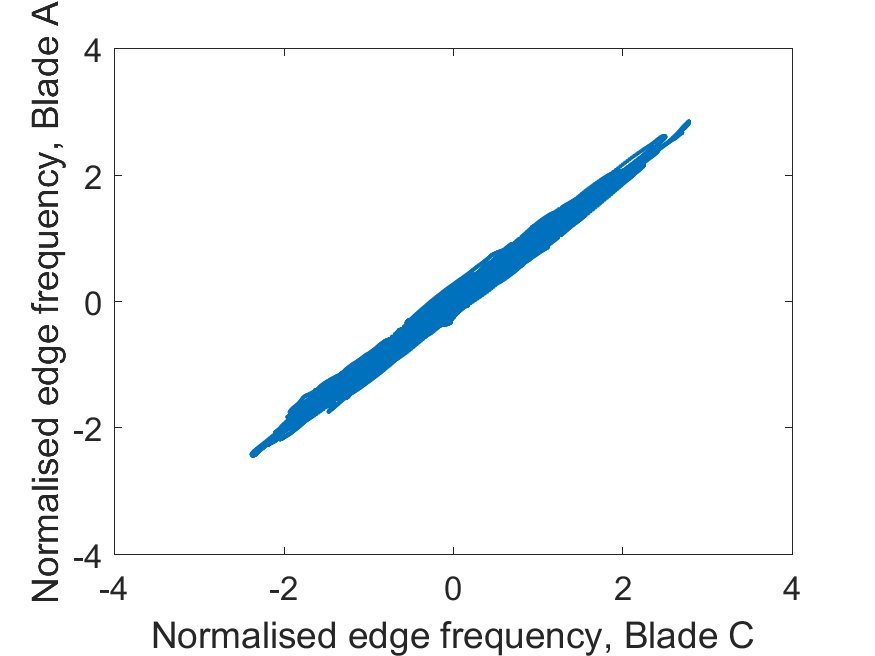}}
    \hspace{0in}
  \vspace{0in}
  \caption{Correlation between normalised edge frequencies for: (a) Blade A vs. Blade B, (b) Blade B vs. Blade C, and (c) Blade C vs. Blade A.}
  \label{fig:freq_one_to_one}
\end{figure}

Next, the various features were analysed to assess whether they contributed to the observed properties of the blade (i.e.\ the edge frequencies). Figure \ref{fig:freq_one_to_one} shows a representative example of how the estimated edge frequencies of each blade correlate to each other. Note that the three blades on the turbine have been labelled A, B, and C.

Based on Figure \ref{fig:freq_one_to_one}, it is clear to see that these frequencies correlate well with each other in a linear fashion. Over time, as the blades age, the correlations generally worsen, and when damage occurs, the relationships generally break down. Therefore, the primary motivation behind the SHM system was to evaluate when the correlations worsened and/or broke down. It should be noted here that for the examples of correlations shown in Figure \ref{fig:freq_one_to_one}, simple methods such as Bayesian linear regression would be sufficient rather than the more complicated GPs. However, as shall be seen later, the correlations between the edge frequencies of each blade during normal condition are typically not as linear and noise-free on other wind turbines (see Figure \ref{fig:freq_one_to_one_temp_siteB}). Since GPs are more robust to such scenarios (since the functional forms need not be predetermined), they are the chosen method for this work.

\begin{figure}[H]
  \vspace{0pt}
  \centering
   \subfloat[]{\includegraphics[scale=0.405, trim=0cm 0cm 0cm 0cm, clip=true]{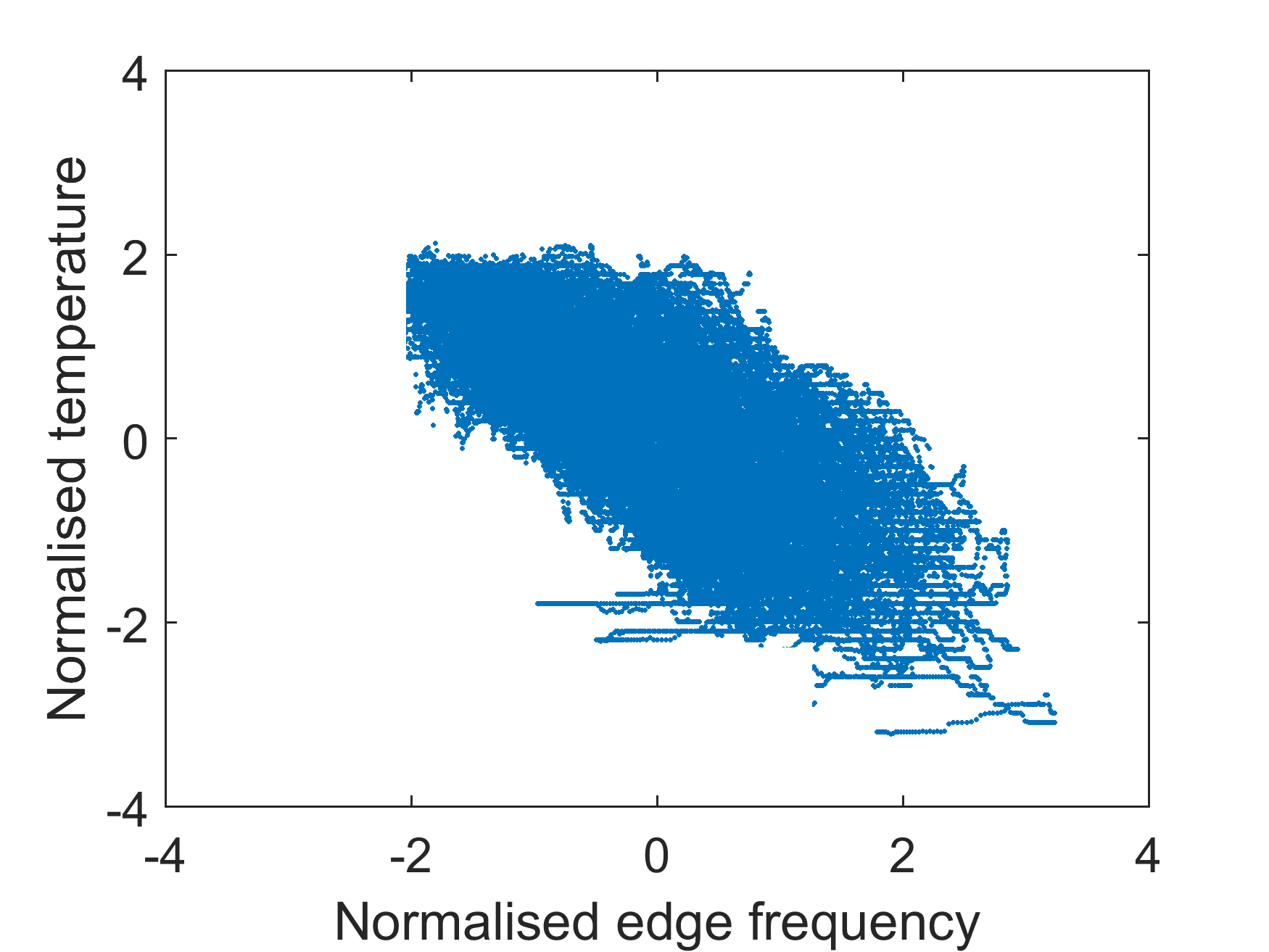}}
  \hspace{0in}
   \subfloat[]{\includegraphics[scale=0.405, trim=0cm 0cm 0cm 0cm, clip=true]{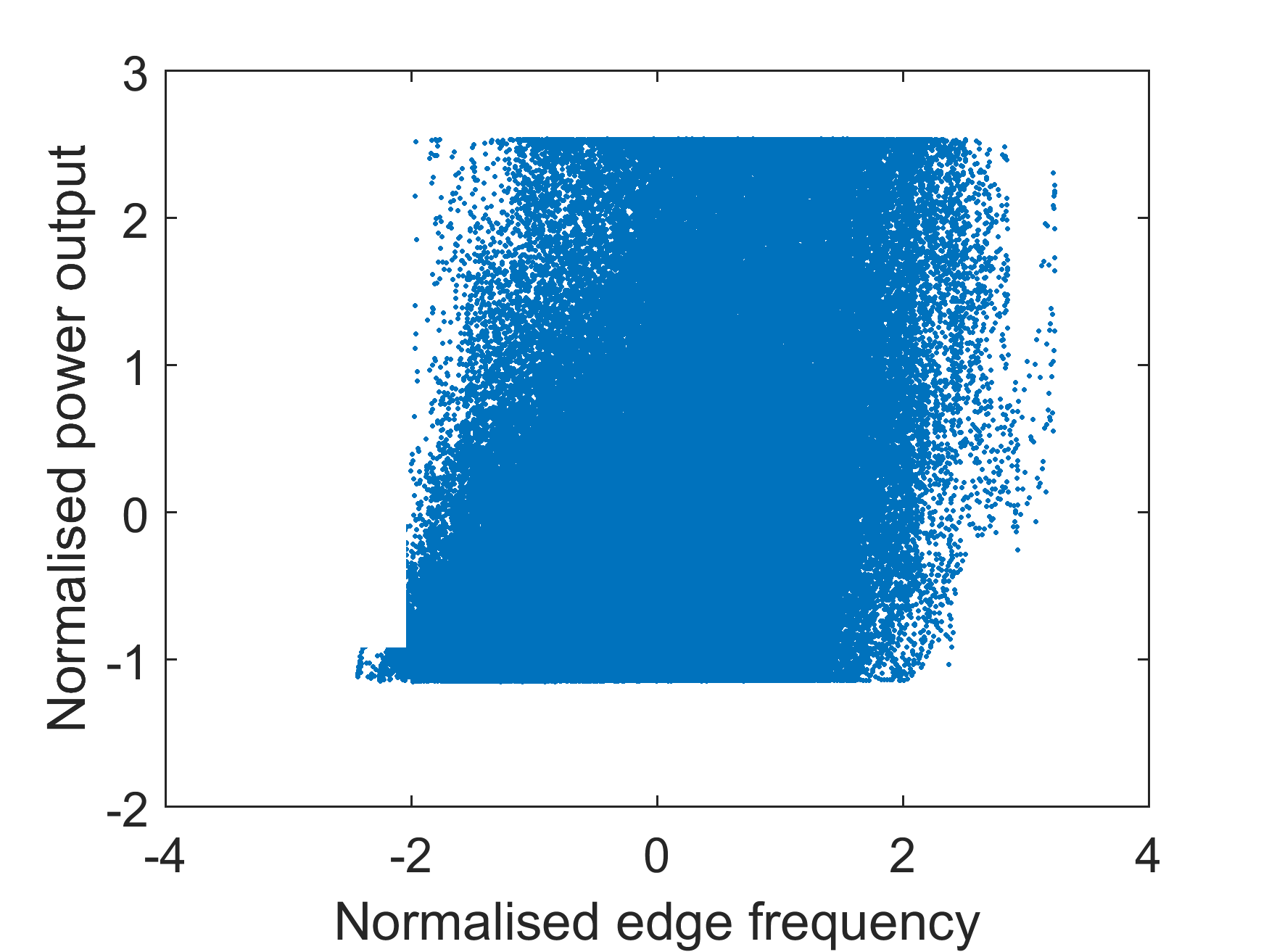}}
    \hspace{0in}   \subfloat[]{\includegraphics[scale=0.405, trim=0cm 0cm 0cm 0cm, clip=true]{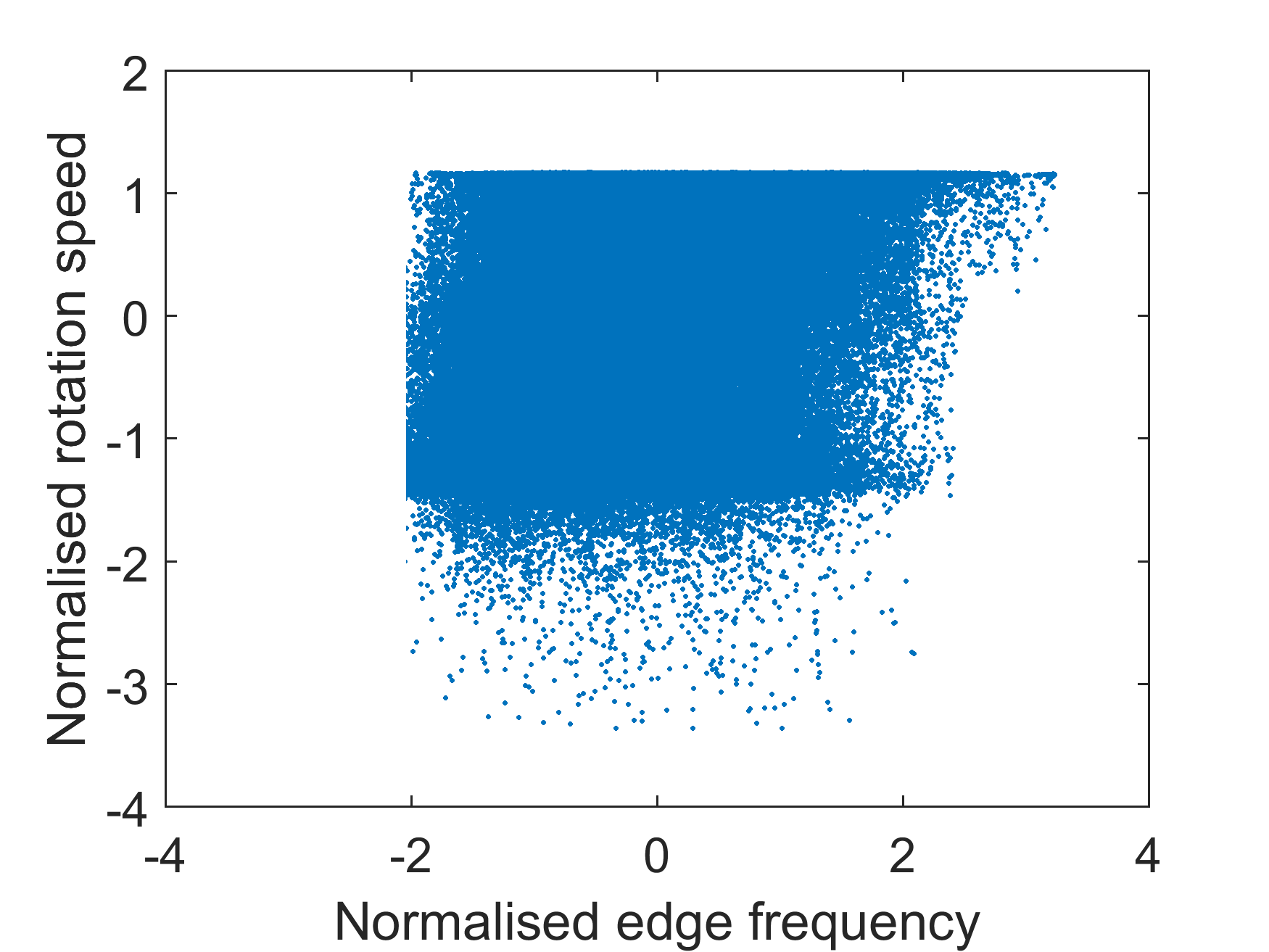}}
    \hspace{0in}
   \subfloat[]{\includegraphics[scale=0.405, trim=0cm 0cm 0cm 0cm, clip=true]{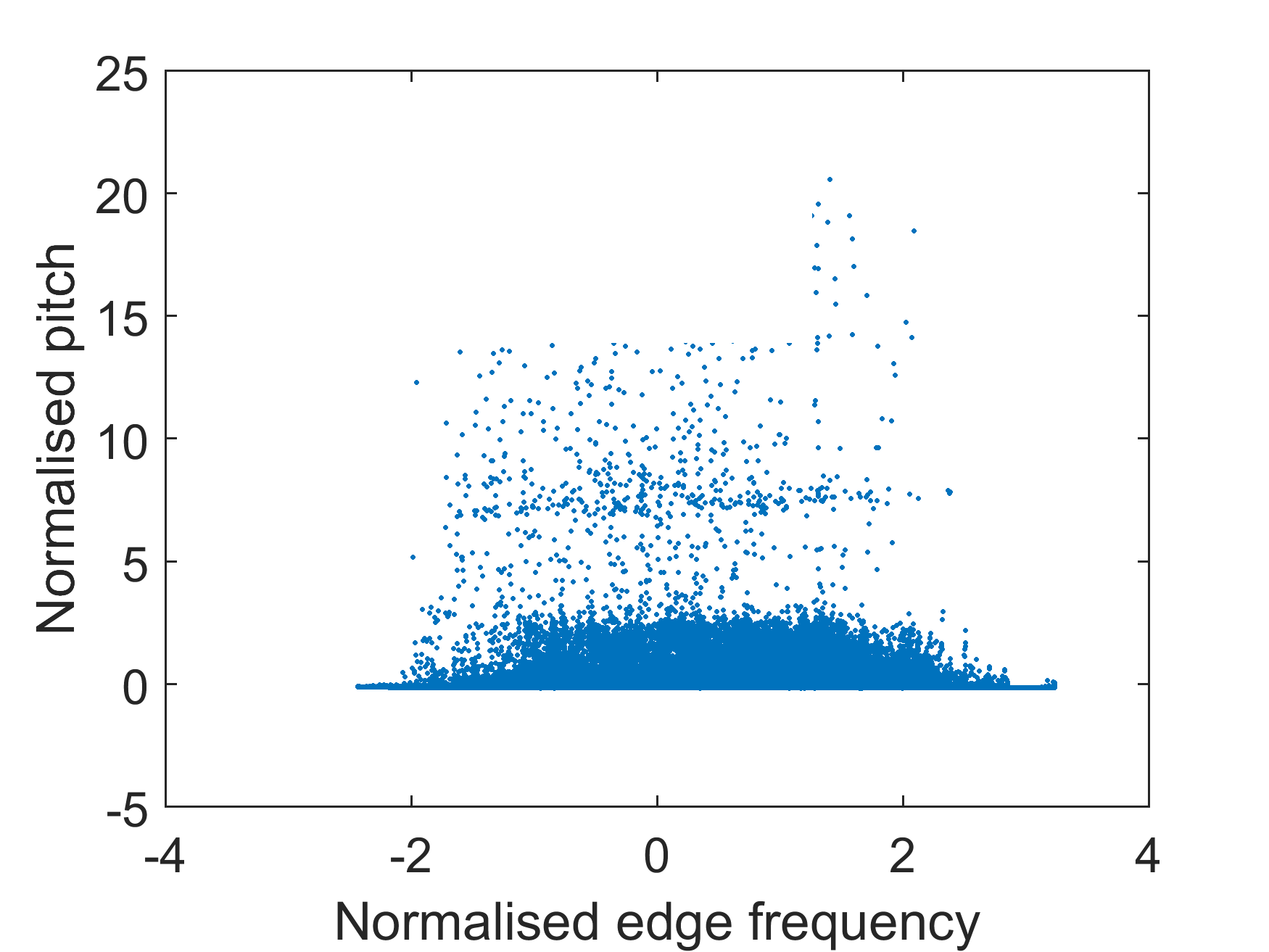}}
    \hspace{0in}
  \vspace{0in}
  \caption{Correlation between normalised edge frequencies and normalised: (a) ambient temperature, (b) power output, (c) generator rotation speed, and (d) pitch angle.}
  \label{fig:EOV_correlation}
\end{figure}

Apart from the study of correlations between the respective edge frequencies, correlations between other EOVs and the edge frequencies of each blade were also studied. Figures \ref{fig:EOV_correlation} (a) to (d) illustrate exemplar relationships between the EOVs (ambient temperature, power output, generator rotation speed, and pitch angle, respectively) with respect to the estimated frequencies.

\begin{figure}[H]
  \vspace{0pt}
  \centering
   \subfloat[]{\includegraphics[scale=0.405, trim=0cm 0cm 0cm 0cm, clip=true]{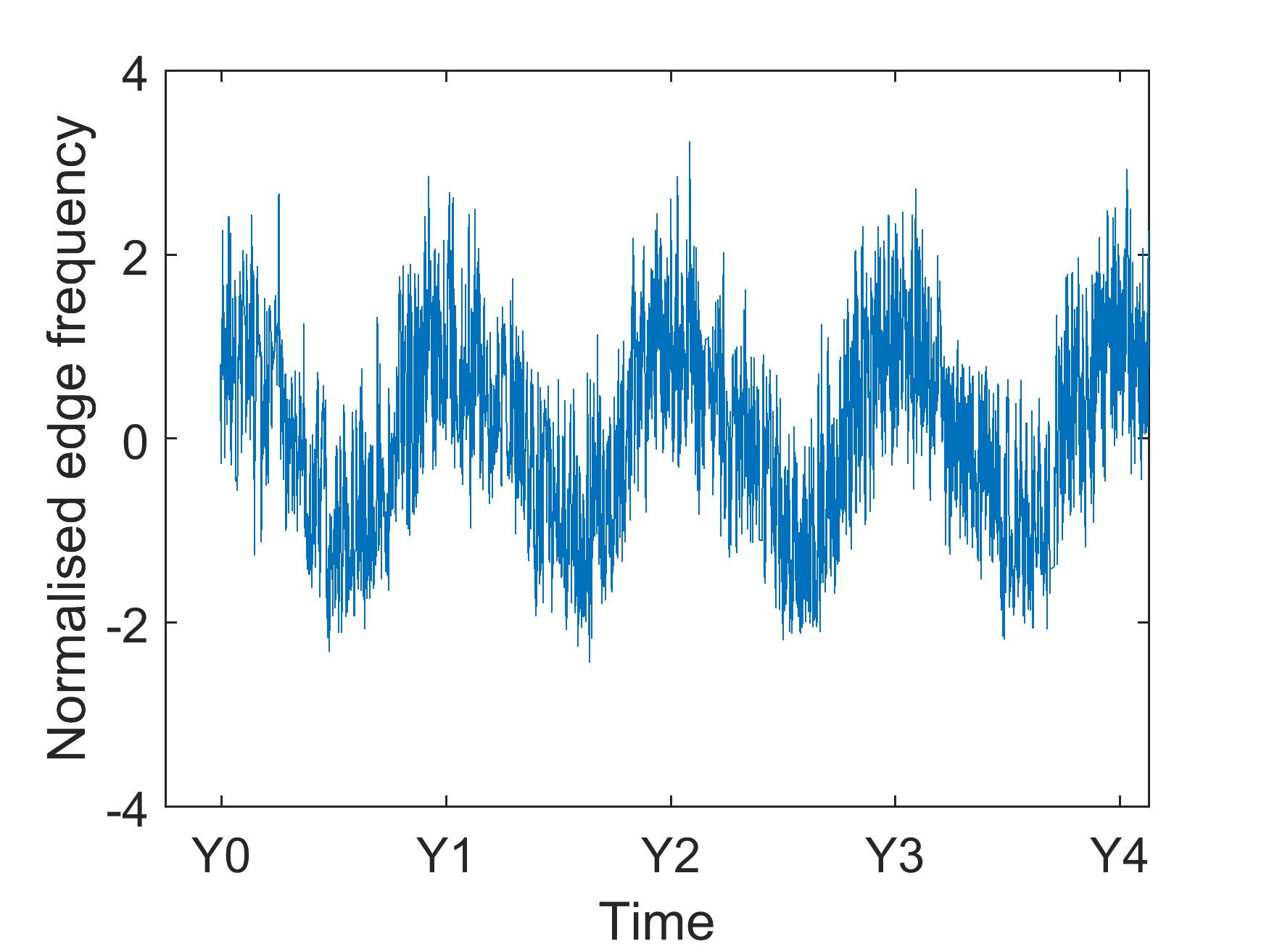}}
  \hspace{0in}
   \subfloat[]{\includegraphics[scale=0.405, trim=0cm 0cm 0cm 0cm, clip=true]{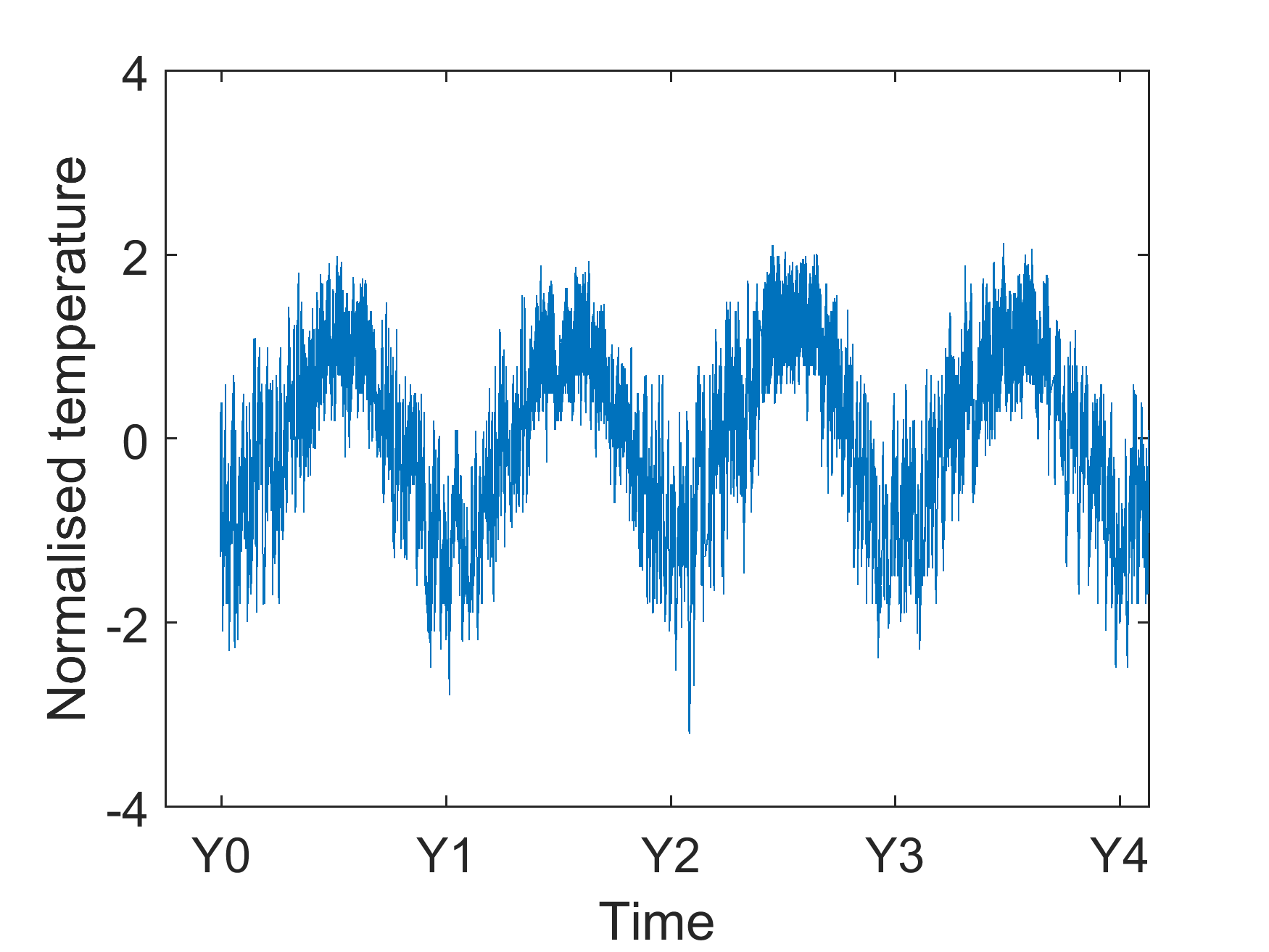}}
    \hspace{0in}
  \vspace{0in}
  \caption{Time series of normalised: (a) frequency, and (b) ambient temperature.}
  \label{fig:cyclic_variations}
\end{figure}

As is indicated in Figure \ref{fig:EOV_correlation}, it was found that, out of the EOVs considered, only the ambient temperature had any systematic influence on the estimated frequencies. When the estimated frequencies and temperatures are viewed over time, it is evident that the cyclical nature of the estimated frequencies occur due to the seasonal variations, as shown in Figure \ref{fig:cyclic_variations}.

The use of these EOVs (other than ambient temperature) in the machine learning tasks that follow were expected to lead to less accurate GP predictions since they do not contribute to the observed behaviour of the edge frequencies. Due to this fact, only the edge frequencies and ambient temperature features were selected for further processing.

In the next section, the theory of Gaussian processes (GPs) is provided.

\subsection{Gaussian Processes}\label{subsec:gp_theory}

Gaussian processes (GPs) implement a nonparametric probabilistic regression. They are best understood in terms of a function space viewpoint \cite{rasmussen2006gaussian}. Let $\boldsymbol{X}$ be an $N \times d$ matrix of inputs $\{\boldsymbol{x}_1,\boldsymbol{x}_2,\cdots,\boldsymbol{x}_N\}$, with $N$ observations and $d$ variables (also referred to as dimensions), $f(\boldsymbol{X})$ denote the underlying functions of the GP, $\{f(\boldsymbol{x}_1), f(\boldsymbol{x}_2),\cdots,f(\boldsymbol{x}_N)\}$, and $\boldsymbol{y}$ be the output vector, which is actually observed, and can be corrupted by noise,

\begin{equation}\label{eq:gp1}
\boldsymbol{y} =f(\boldsymbol{X}) + \boldsymbol{\varepsilon}
\end{equation} 

\noindent where $\boldsymbol{\varepsilon}$ is a zero mean Gaussian noise process, with variance ${\sigma_N}^2$. Note that for the sake of conciseness, $f(\boldsymbol{X})$ will henceforth simply be denoted as $\boldsymbol{f}$.

In the Bayesian approach, the prior belief of the form of the GP, $\boldsymbol{f}$ is utilised along with the likelihood function of the observed data to calculate its posterior distribution. This can be expressed as,

\begin{equation}\label{eq:bayes}
\text{posterior} = \frac{\text{likelihood} \times \text{prior}}{\text{marginal likelihood}}
\end{equation} 

\begin{equation}\label{eq:bayes2}
p(\boldsymbol{f}|\boldsymbol{y},\boldsymbol{X})  = \frac{p(\boldsymbol{y}|\boldsymbol{X},\boldsymbol{f})  p(\boldsymbol{f}|\boldsymbol{X})} {p(\boldsymbol{y}|\boldsymbol{X})} = \frac{p(\boldsymbol{y}|\boldsymbol{X},\boldsymbol{f})  p(\boldsymbol{f}|\boldsymbol{X})}{\int{p(\boldsymbol{y}|\boldsymbol{X},\boldsymbol{f})  p(\boldsymbol{f}|\boldsymbol{X}) d\boldsymbol{f}}}
\end{equation}

\noindent where $p(\boldsymbol{f}|\boldsymbol{y},\boldsymbol{X})$ is the posterior distribution (probability of functions given the outputs and inputs, $p(\boldsymbol{y}|\boldsymbol{X},\boldsymbol{f})$ is the likelihood function (probability of outputs, given inputs and functions), $p(\boldsymbol{f}|\boldsymbol{X})$ is the prior distribution (probability of the functions, given inputs), and $p(\boldsymbol{y}|\boldsymbol{X})$ is the marginal distribution (probability of outputs, given only the inputs). $p(\boldsymbol{y}|\boldsymbol{X})$ is independent of the functions because it is integrated over the entire functional space of $\boldsymbol{f}$. 

In the framework of GPs, functions between inputs and outputs can be completely defined by a mean function, $m(\boldsymbol{x})$, and a covariance function, $k(\boldsymbol{x}_p,\boldsymbol{x}_q)$, and these are in turn defined by hyperparameters \cite{rasmussen2006gaussian}. Note that the subscripts are used to distinguish one input vector from another. This definition is analogous to a Gaussian distribution, which can be fully described using a mean and a variance. GPs therefore provide a distribution of functions between inputs and outputs. The mean and covariance functions are expressed as,

\begin{equation} \label{eq:mean_fun}
m(\boldsymbol{x}) = E[f(\boldsymbol{x})]
\end{equation}

\begin{equation}\label{eq:cov_fun}
k(\boldsymbol{x}_p,\boldsymbol{x}_q) = E[(f(\boldsymbol{x}_p) - m(\boldsymbol{x}_p))(f(\boldsymbol{x}_q) - m(\boldsymbol{x}_q))]
\end{equation} 

\noindent The covariance function is a vital ingredient of the GP, as it characterises the general properties of the GP. Typically, when $\boldsymbol{x}_p$ and $\boldsymbol{x}_q$ are almost identical, $k(\boldsymbol{x}_p,\boldsymbol{x}_q) $   would yield high values (high correlation), and as they grow in distance, the values reduce in magnitude (low correlation). Due to this, when predictions are performed, new inputs that geometrically lie close to previously observed inputs will have a greater influence than those that lie far away.

\subsubsection{Covariance functions}
The choice of covariance functions is one of fundamental importance. Depending on the problem being solved, there are numerous covariance functions (also referred to as kernels in the literature) available to use. Examples include squared-exponential, Mat\'ern, $\gamma$-exponential, rational quadratic, and the Bayesian linear covariance functions \cite{rasmussen2006gaussian}. These functions can be combined using basic algebra (addition, multiplication, etc.) to form advanced covariance functions. In the work presented in this article, only the squared-exponential and Bayesian linear covariance functions are used, and hence the discussion will be focussed on these.

The squared-exponential covariance function is defined as,

\begin{equation} \label{eq:sqexp_cov}
k_{SE}(\boldsymbol{x}_p,\boldsymbol{x}_q) = {\sigma_f}^2 \exp\bigg(-\frac{(\boldsymbol{x}_p - \boldsymbol{x}_q)^T(\boldsymbol{x}_p - \boldsymbol{x}_q)}{2\lambda^2}\bigg)
\end{equation}

\noindent where ${\sigma_f}^2$ is a hyperparameter that represents the signal variance, and $\lambda$ is a length-scale hyperparameter (which controls how smooth the function appears). Note that $(\boldsymbol{x}_p - \boldsymbol{x}_q)^T(\boldsymbol{x}_p - \boldsymbol{x}_q)$ is simply the squared Euclidean distance between input points.

The Bayesian linear covariance function is defined as,

\begin{equation}\label{eq:lin_cov}
k_{BL}(\boldsymbol{x}_p,\boldsymbol{x}_q) =  {\sigma_0}^2\boldsymbol{x}_p\cdot\boldsymbol{x}_q
\end{equation} 

\noindent where $ {\sigma_0}^2$ is also a hyperparameter that represents the signal variance. This covariance function is classed as a \textit{dot-product covariance function}, for obvious reasons.

Finally, input noise can also be taken into account, i.e.\ when observations are noisy. This is defined via,

\begin{equation} \label{eq:lin_cov2}
k_{N}(\boldsymbol{x}_p,\boldsymbol{x}_q) =  {\sigma_N}^2 \delta(\boldsymbol{x}_p,\boldsymbol{x}_q) = {\sigma_N}^2\boldsymbol{I}
\end{equation}

\noindent where ${\sigma_N}^2$ is a hyperparameter that represents the variance of the noise process and $\delta$ is the Kronecker-delta function, which can simply be viewed as the identity matrix, $\boldsymbol{I}$.

For the GPs associated with this work, a sum of the above covariance functions was considered to explain the observed data. There are three distinct covariance matrices associated with the predictive GPs:

\begin{enumerate}
	\item $\boldsymbol{K_\theta(X,X)}$: Covariance matrix between training data only. For the sake of convenience, this will simply be denoted $\boldsymbol{K_\theta}$.
	\begin{equation} \label{eq:cov_train}
	\boldsymbol{K_\theta} = k_{SE} (\boldsymbol{x}_p,\boldsymbol{x}_q) +  k_{BL} (\boldsymbol{x}_p,\boldsymbol{x}_q) +  k_N (\boldsymbol{x}_p,\boldsymbol{x}_q)
	\end{equation}
	
	\item $\boldsymbol{K_* (X,X^* )}$: Cross covariance matrix between training and test data. The asterisk superscript denotes test data. Note that $\boldsymbol{K_* (X^*,X)}$ can also be calculated separately, but it is simply the transpose of $\boldsymbol{K_* (X,X^* )}$. For conciseness, $\boldsymbol{K_* (X,X^* )}$ will be denoted $\boldsymbol{K_*}$.
	
	\begin{equation} \label{eq:cov_testtrain}
	\boldsymbol{K_*} = k_{SE} (\boldsymbol{x}_p,\boldsymbol{x}_{q}^{\boldsymbol{*}}) +  k_{BL} (\boldsymbol{x}_p,\boldsymbol{x}_{q}^{\boldsymbol{*}}) 
	\end{equation} 
	
	\item $\boldsymbol{K_{**} (X^*,X^* )}$: Covariance matrix between test data only. This will simply be denoted $\boldsymbol{K_{**}}$.
	
	\begin{equation}\label{eq:cov_test}
	\boldsymbol{K_{**}}= k_{SE} (\boldsymbol{x}_{p}^{\boldsymbol{*}},\boldsymbol{x}_{q}^{\boldsymbol{*}}) +  k_{BL} (\boldsymbol{x}_{p}^{\boldsymbol{*}},\boldsymbol{x}_{q}^{\boldsymbol{*}}) 
	\end{equation} 
\end{enumerate}

It should be noted that in the latter two covariance functions, the noise term is not added for the prediction of the noise-free underlying function, $\boldsymbol{f^*}$. 

Now that the background of GPs has been introduced, it is necessary to discuss how they can be used in regression problems. Matters will be elaborated using the terms in the Bayes\rq{} theorem: the prior, the likelihood, and the marginal likelihood. The marginal likelihood is utilised in the training process, where the hyperparameters that define the GP are learnt.

\subsubsection{The prior distribution}
The prior to the functions, $\boldsymbol{f}$ has the form,

\begin{equation}\label{eq:prior1}
\boldsymbol{f} \sim \mathcal{GP}(0,\boldsymbol{K})
\end{equation} 

\noindent where the 0 indicates a zero-mean function. $\boldsymbol{K}$ is now the covariance matrix, without the noise term. The benefit of incorporating a mean function is that in the absence of training data in a given region of the input space, the prediction of the GP will approach the mean trend that is defined by the mean function. This is because it specifies a belief in the relationship between inputs and outputs having a certain functional form. It is noteworthy that it is not necessary to limit the form of the GP using a mean. This is because, in practice, data is typically de-trended, and so, mean trends are already removed \cite{murphy2012machine}. In any case, the prior distribution gets updated in each step of the training process, and so the mean function can be set to zero. However, this does not imply that the output of the GP will have zero mean, and is hence not a limiting issue. It has a distribution,

\begin{equation}\label{eq:prior2}
p(\boldsymbol{f}|\boldsymbol{X})= \mathcal{N}(0,\boldsymbol{K})
\end{equation} 

\subsubsection{The likelihood distribution}\label{subsec:likelihood}

$\boldsymbol{f}$ is a realisation of the underlying GP with inputs, $\boldsymbol{X}$. Because $\boldsymbol{y}$ is simply $\boldsymbol{f}$  with additive Gaussian noise ($\mathcal{N}(0, {\sigma_N}^2)$), the likelihood distribution can be stated as,

\begin{equation} \label{eq:likelihood}
p(\boldsymbol{y}|\boldsymbol{X},\boldsymbol{f}) = \mathcal{N}(\boldsymbol{f}, {\sigma_N}^2\boldsymbol{I})
\end{equation}

\subsubsection{The marginal likelihood and hyperparameter learning} \label{subsec:marginal_likelihood}

The denominator in equation (\ref{eq:bayes2}) is important in estimating (or learning) the hyperparameters. It is also normally distributed with the form,

\begin{equation} \label{eq:marg_likelihood}
p(\boldsymbol{y}|\boldsymbol{X}) = \mathcal{N}(0, \boldsymbol{K_\theta})
\end{equation}

\noindent where $\boldsymbol{K_\theta}$ is the covariance matrix with additive noise, i.e.\ $\boldsymbol{K_\theta} = \boldsymbol{K} + {\sigma_N}^2 \boldsymbol{I}$. The marginal likelihood is usually expressed in terms of its logarithmic transformation: the log marginal likelihood \cite{rasmussen2006gaussian},

\begin{equation} \label{eq:log_marg_likelihood}
\mathcal{L} = \log{}p(\boldsymbol{y}|\boldsymbol{X})= -\frac{1}{2} \boldsymbol{y}^T \boldsymbol{K_\theta}^{-1}\boldsymbol{y} - \frac{1}{2} \log{}|\boldsymbol{K_\theta}| - \frac{N}{2} \log{}(2\pi)
\end{equation}

\noindent where $N$ is the number of training points, and the other definitions are consistent with the preceding discussion. When the hyperparameters of the covariance functions are being learnt, the goal is to maximise the marginal likelihood with respect to the hyperparameters. Alternatively, the negative log marginal likelihood can also be minimised. If the hyperparameters of the covariance functions can be combined in a tuple, $\boldsymbol{\varphi} = \{\sigma_f,\lambda,\sigma_0, \sigma_N\}$, the partial derivative of the negative log marginal likelihood can be expressed as,

\begin{equation} \label{eq:pd_nlml}
\frac{\partial}{\partial\boldsymbol{\varphi}_i} (-\mathcal{L})=  \frac{1}{2} tr\bigg[\boldsymbol{K_\theta}^{-1} \frac{\partial\boldsymbol{K_\theta}}{\partial\boldsymbol{\varphi}_i}\bigg] - \frac{1}{2}\boldsymbol{y}^T\boldsymbol{K_\theta}^{-1} \frac{\partial\boldsymbol{K_\theta}}{\partial\boldsymbol{\varphi}_i}\boldsymbol{K_\theta}^{-1} \boldsymbol{y}
\end{equation}

\noindent where $tr$ is the trace operator (i.e.\ sum of the diagonal elements of the matrix). In practice, the negative log marginal likelihood of GPs may have several local minima. Therefore, solving this problem becomes one of numerical optimisation. There are several routes one can take, for example gradient descent, differential evolution, Broyden-Fletcher-Goldfarb-Shanno algorithm, etc. There are no right or wrong choices in the optimisation routines, but each optimiser can have advantages and disadvantages associated with them. In this work, the Nelder-Mead simplex algorithm was used as the optimiser \cite{nelder1965simplex}.

\subsubsection{Posterior distribution} \label{subsec:posterior}

The posterior distribution can be obtained by combining equations (\ref{eq:prior2}), (\ref{eq:likelihood}) and (\ref{eq:marg_likelihood}), and using Bayes\rq{} theorem - equation (\ref{eq:bayes}); this is given by,

\begin{equation} \label{eq:posterior}
p(\boldsymbol{f}|\boldsymbol{X},\boldsymbol{y}) = \mathcal{N}(\boldsymbol{K}^T\boldsymbol{K_\theta}^{-1}\boldsymbol{y}, \boldsymbol{K}-\boldsymbol{K}^T\boldsymbol{K_\theta}^{-1}\boldsymbol{K})
\end{equation}

\subsubsection{Predictions using GPs}\label{subsec:predictions}

Thus far, the necessary ingredients of the GPs have been shown. The next task is to use these for the main goal of the GPs: prediction. Let $\boldsymbol{x^*}$ denote a test input vector, (i.e.\ a matrix which is now going to be used to predict the outputs). The corresponding noise-free test output is $\boldsymbol{f^*}$. Now, the observed (noisy) training outputs and the noise-free test outputs are jointly Gaussian distributed according to the prior distribution, and can be expressed as,

\begin{equation} \label{eq:predictive_distribution}
\begin{bmatrix}
\boldsymbol{y} \\
\boldsymbol{f^*}
\end{bmatrix}
=
\mathcal{N}
\begin{pmatrix}
0,
\begin{bmatrix}
\boldsymbol{K_\theta} & \boldsymbol{K_*} \\
\boldsymbol{K_*}^T & \boldsymbol{K_{**}}
\end{bmatrix}
\end{pmatrix}
\end{equation}

\noindent where the respective covariance matrices were defined in equations (\ref{eq:cov_train}), (\ref{eq:cov_testtrain}), and (\ref{eq:cov_test}). The corresponding conditional distribution can be expressed as,

\begin{equation} \label{eq:conditional_distribution}
p(\boldsymbol{f^*}|\boldsymbol{X},\boldsymbol{y},\boldsymbol{x^*}) = \mathcal{N}(\boldsymbol{\bar{f^*}}, cov(\boldsymbol{f^*}))
\end{equation}

\noindent where the predictive mean, $\boldsymbol{\bar{f^*}}$ of the GP can be calculated using,

\begin{equation} \label{eq:predictive_mean}
\boldsymbol{\bar{f^*}} = \boldsymbol{K_*}^T\boldsymbol{K_\theta}^{-1}\boldsymbol{y}
\end{equation}

\noindent and the predictive covariance, $cov(\boldsymbol{f^*})$ of the GP can be calculated using, 

\begin{equation} \label{eq:predictive_cov}
cov(\boldsymbol{f^*})=\boldsymbol{K_{**}} - \boldsymbol{K_*}^T \boldsymbol{K_\theta}^{-1} \boldsymbol{K_*}
\end{equation}

\noindent The diagonal of the matrix given by equation (\ref{eq:predictive_cov}) is then used to give the uncertainties (in terms of variance) in prediction. The noise variance can be added to the predictive variance to take uncertainty related to noisy test outputs into consideration, and hence instead of finding the predictive distribution of $\boldsymbol{f^*}$, the predictive distribution of $\boldsymbol{y^*}$ can be calculated. Note that this does not affect the predictive mean, only the predictive covariance.

\subsection{Application of machine learning}

\begin{figure}[H]
\vspace{0pt}
\centering
\includegraphics[scale = 0.35, trim=9.5cm 1cm 9.5cm 0cm, clip=true]{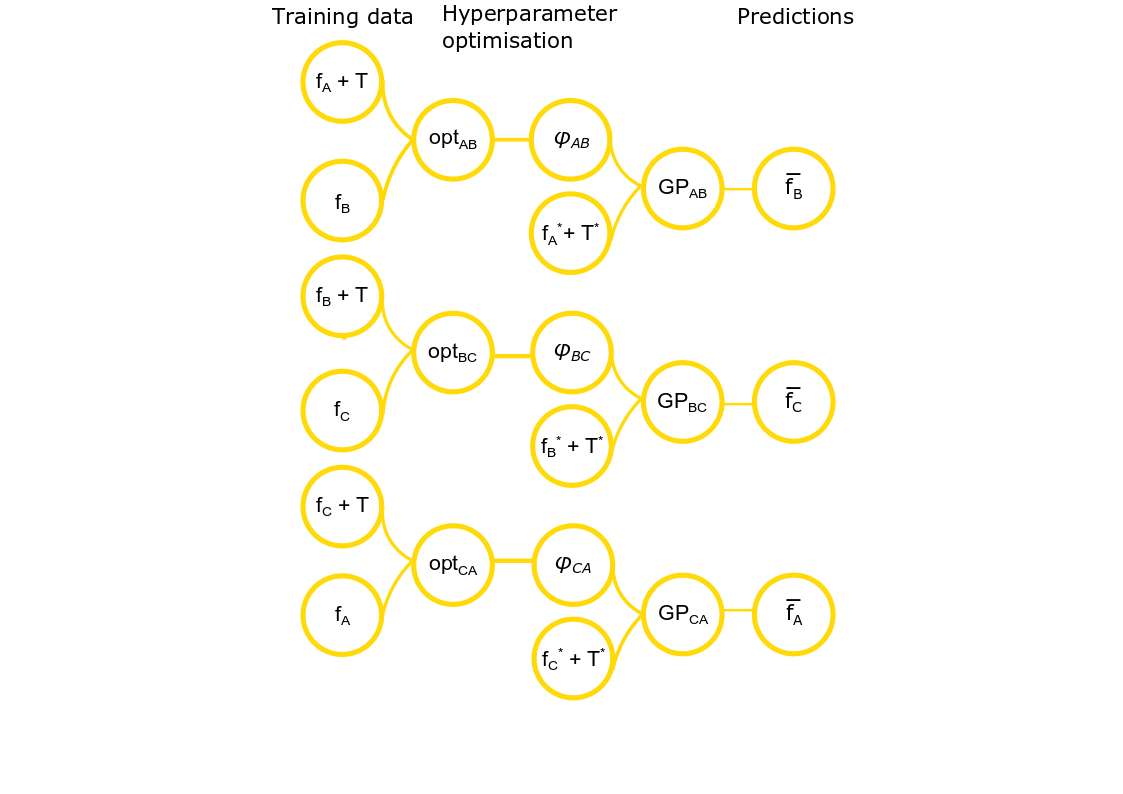}
\centering
\caption{Flow of data in the proposed methodology.}
\label{fig:machine_learning}
\end{figure}

Figure \ref{fig:machine_learning} provides an overview of the flow of data in the machine learning phase of the SHM scheme, which includes training, optimisation, and Gaussian Process (GP) predictions.

Figure \ref{fig:machine_learning} shows that there are a total of three GPs that perform the predictions. However, before any predictions could be performed, the first task was to learn the hyperparameters, $\varphi$ of the models that controlled their predictive capabilities - and this entailed a training process. Here, a subset of the selected feature data were selected for the training. 

When choosing the training data, it was important to consider only data when the blades were in a normal (healthy) condition. However, the conditions of the blades were not known beforehand. It was therefore decided that training data would only be selected from the first two years of the turbine operation, assuming that the blades were in a normal condition throughout this period. Within this period of two years, 2500 data points were randomly selected from a uniform distribution. Choosing this two year time period for training ensured that the selected data spanned the entire normal operating range of the blades. The random selection ensured that there was no bias to any specific operating condition in the training phase. It should be noted that more training data would increase the computational costs of the GPs, since computational costs are $\mathcal{O}(N^3)$, where $N$ is the number of training points. There are sparse methods available to reduce the computational load associated with training, but these were not employed in this work.

The estimated frequencies, $f$ of one blade (e.g.\ for blade A, $f_A$) were combined with the ambient temperatures, $T$ to form a matrix of the training inputs. Referring to Section \ref{subsec:gp_theory}, this is the matrix, $\boldsymbol{X}$. The estimated frequencies of another blade (e.g.\ for blade B, $f_B$) formed the training outputs. This is the vector, $\boldsymbol{y}$. in Section \ref{subsec:gp_theory}. These training data were used to learn the hyperparameters ($\varphi$) for the respective GPs via the Nelder-Mead method. During the learning phase, the covariance matrices, equations (\ref{eq:cov_train} - \ref{eq:cov_test}) were evaluated several times, which is why it was stated that the computational costs of GPs increases significantly with the number of training points. The hyperparameters selected corresponded to those that yielded the largest maximum log marginal likelihood, equation (\ref{eq:log_marg_likelihood}). The Nelder-Mead algorithm identifies this by minimising the negative log marginal likelihood, i.e.\ the negative of equation (\ref{eq:log_marg_likelihood}).  These hyperparameters control the generalisation properties of the GPs, i.e.\ how the GPs adapt to the nature of the problem being tested. The asterisk symbols, $^*$ shown in Figure \ref{fig:machine_learning} are used to denote test data, i.e.\ data not included in the training set. The test data,  $f^*$ and $T^*$ were combined in a matrix (referred to as the matrix,  $\boldsymbol{X^*}$ in Section \ref{subsec:gp_theory}), and were then used in the various GPs (GP$_{AB}$, GP$_{BC}$, and GP$_{CA}$), using the respective hyperparameters ($\varphi_{AB}$, $\varphi_{BC}$, and $\varphi_{CA}$), to evaluate the predicted frequencies, $\bar{f_B}$, $\bar{f_C}$, and $\bar{f_A}$, respectively. The bar symbol  ( $\bar{}$ )  indicates a prediction, rather than an actual estimate of the frequencies. The predictions are referred to as $\boldsymbol{f^*}$ in Section \ref{subsec:gp_theory}, and are calculated using equation (\ref{eq:predictive_mean}). To elucidate,

\begin{itemize}
\item in GP$_{AB}$, ${f_A}^*$ and $T^*$ were used to predict $\bar{f_B}$ using hyperparameters $\varphi_{AB}$,
\item in GP$_{BC}$, ${f_B}^*$ and $T^*$ were used to predict $\bar{f_C}$ using hyperparameters $\varphi_{BC}$, and
\item in GP$_{CA}$, ${f_C}^*$ and $T^*$ were used to predict $\bar{f_A}$ using hyperparameters $\varphi_{CA}$.
\end{itemize}

\subsection{Diagnostics and actions}

\begin{figure}[H]
\vspace{0pt}
\centering
\includegraphics[scale = 0.35, trim=5cm 9cm 5cm 6cm, clip=true]{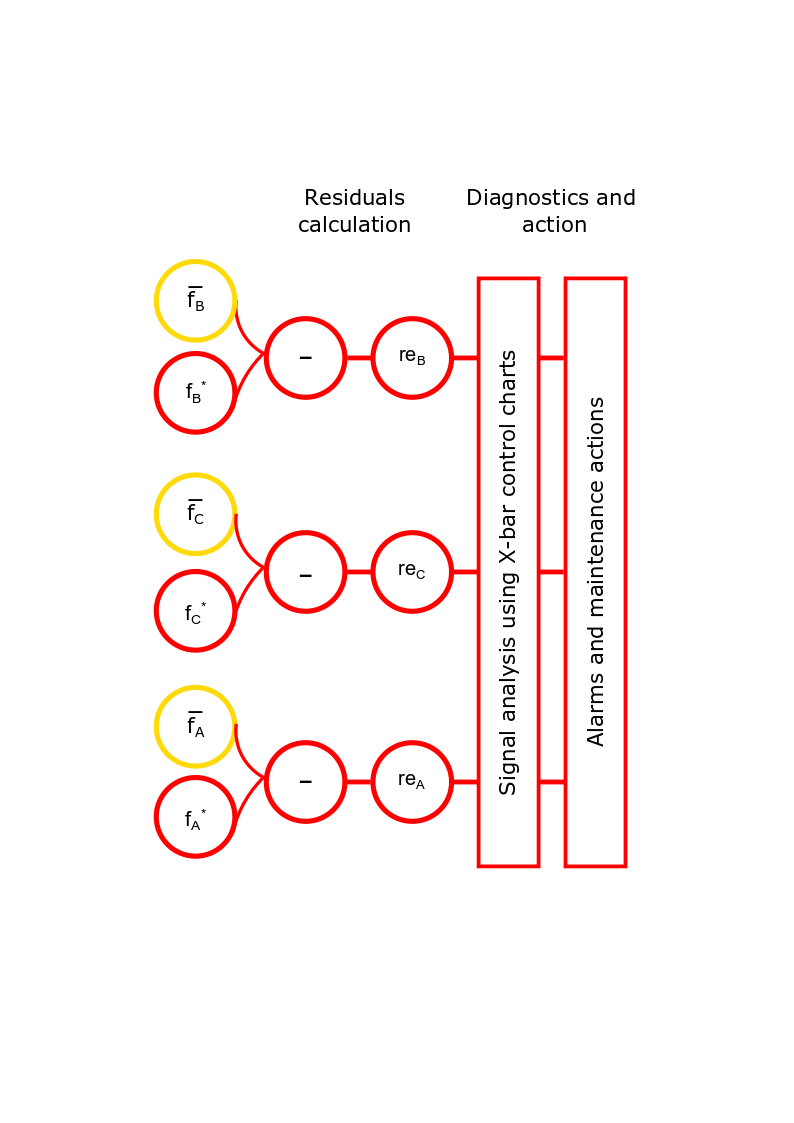}
\centering
\caption{Diagnostics of GP residuals.}
\label{fig:diagnostics_and_action}
\end{figure}

Figure \ref{fig:diagnostics_and_action} shows the tasks involved in the diagnostics and actions section. 

Following the frequency predictions for each blade via the respective GPs, residual errors were calculated between the predicted frequencies and the actual estimations (for example, the GP residual error for blade B, $re_B$ was defined as the difference between $\bar{f_B}$ and ${f_B}^*$).

The assessment of whether the blades were healthy or not was performed using X-bar control charts. In the X-bar control charts, averages of the GP residuals were taken at regular time intervals. For the purposes of this work, averages were taken regularly over a 28 day interval. During a brief period that followed the GP training phase detailed in the previous section, 3$\sigma$ thresholds were also calculated. These were evaluated over a period of 6 months. The calculated averages of the X-bar control chart were then compared with the 3$\sigma$ thresholds, and if the averages exceeded the thresholds, it was concluded that the blades were no longer in a normal condition.

During the development of this SHM system, maintenance actions were not taken. In practice, once the GP residuals surpass the thresholds of the X-bar control charts, warnings would be generated before alarms occur. In such a scenario, maintenance technicians would be sent to inspect the blades, they would assess the level of damage, and finally would decide whether to repair or replace the turbines.

\subsection{Summary} \label{sec:summary}

The following summarises the SHM system as detailed in the preceding sections.

\begin{enumerate}
\item Data acquisition.
\item Data cleansing (removal of outliers).
\item Feature selection.
\item GP training (identification of hyperparameters).
\item GP prediction.
\item Calculation of residual errors.
\item X-bar control chart analysis of GP residuals.
\end{enumerate}

\section{Case studies}

Now that the diagnostic methodology and theory have been described in detail, it is important to illustrate its performance. Prior to the results and discussion of the case studies, the detailed steps of the employed machine learning methodology are stated.

\subsection{Detailed implementation}

In this section, a detailed description is given/reiterated regarding the implementation of the machine learning methods described thus far, referring to the seven steps in Section \ref{sec:summary}. That is,

\begin{enumerate}
	\item Data acquisition: In the following sections, three sets of data are introduced. The data are either synthesised or acquired directly from operational wind turbines. In the synthesised case, it is acknowledged that the data does not reflect the true behaviour of blades. However, the method of the data synthesis is described in detail to allow readers to reproduce them and apply the machine learning methodology. In the two cases where the data is acquired from operational wind turbines, a blade on each turbine shown was known to be damaged, and as a result, it either needed to be repaired or replaced. Damage can occur at various locations across the length of the blade, and the types of damage in blades are predominantly cracks or delaminations. In some cases, both occur at the same time. Due to confidentiality issues, however, the specific damage types and locations are not disclosed here. 
	
	\item Data cleansing (removal of outliers): In the case of the synthesised data, outliers are not included in the data, hence this process is not conducted. In the case of the real data, the standard deviations of the dataset are calculated, and data points outside 3$\sigma$ of the data are iteratively removed. This process is repeated until the difference in standard deviation between subsequent iterations is less than 0.1. This value is empirically identified to ensure that outliers due to blade damage are not removed. Note that when an outlier is removed from one of the data channels, data from the other channels with the same timestamp (which may not necessarily be outliers) are also removed.

	\item Feature selection: The main features of interest are the edge frequencies of the blades. It is shown in Example 3 below that the inclusion of temperature as a feature improves the performance of the GPs.
	
	\item GP training (identification of hyperparameters): For each example shown, 2500 training points are selected from a period when the blades on the turbine are considered \lq{healthy}\rq{}. In the case of data from the real turbines, this was chosen to be the first two years of turbine operation. The training points, indexed by the data point numbers, are sampled uniformly via that index. Standard normalisation, equation (\ref{eq:data_norm}) is applied to the entire dataset using the mean and standard deviation values of the training data points. Next, the training input and output edge frequency data are prepared into three pairs (that is, $f_A$ and $f_B$, $f_B$ and $f_C$, and $f_C$ and $f_A$; temperature, $T$ is also included for the training inputs). The hyperparameters, $\phi_{AB}$, $\phi_{BC}$, and $\phi_{CA}$, are then identified by minimising the negative log marginal likelihood, which is equivalent to maximising log marginal likelihood, equation (\ref{eq:log_marg_likelihood}). The Nelder-Mead algorithm is used to accomplish this. 
	\item GP prediction: The identified hyperparameters for each GP are used with the rest of the edge frequency and temperature data (${f_A}^*$, ${f_B}^*$, ${f_C}^*$, and $T^*$) are used to evaluate the predicted edge frequencies, $\bar{f_A}$, $\bar{f_B}$, and $\bar{f_C}$ using equation (\ref{eq:predictive_mean}). Note that for relatively large datasets such as those used in this work, computer memory can easily be used up since the covariance matrices would be too large. For this reason, the GP predictions are performed in smaller chunks (1000 points at a time in this case).
	
	\item Calculation of residual errors: The GP residuals are calculated by subtracting 
	the actual edge frequency data from the predicted edge frequencies. That is,
	
	\begin{equation}
		\begin{aligned}
			re_A &= \bar{f_A} - {f_A}^* \\
			re_B &= \bar{f_B} - {f_B}^* \\
			re_C &= \bar{f_C} - {f_C}^*
		\end{aligned}
	\end{equation}
	
	\item X-bar control chart analysis of GP residuals: In this step, the X-bar control charts of the residuals are calculated. This is performed by calculating averages of the residual errors over periods spanning 28 days. As the GPs are trained over the first two year period on the examples shown from real wind turbines, it is important that the thresholds for the X-bar control chart analyses are not evaluated over the same period. This is because, generally, residual errors are low over training data since optimisation routines aim to minimise these errors as much as possible when identifying the hyperparameters. Thus, such practices can potentially lead to stringent thresholds that may increase false alarm rates. For this reason, a six-month period immediately after the initial two-year training period is chosen. As the standard deviation of the errors during this period may be occasionally large, the following technique is used to calculate robust thresholds. For each of the residual error datasets over this six-month period, small subsets are randomly selected several times (in this work, 20 times). For each subset $i$, the mean $\mu_i$ and standard deviation $\sigma_i$ values are calculated. Finally, averages of the mean and standard deviation data are calculated,  $\bar{\mu}$ and $\bar{\sigma}$, respectively. The thresholds, $thr$ are then calculated using,
	
	\begin{equation}
	thr = \bar{\mu} \pm 3\bar{\sigma}
	\end{equation}
	
	The residual errors for each blade are then compared to the upper and lower thresholds for each blade. Once the residual errors exceed these thresholds, the blade is assumed to be damaged. 
	 
\end{enumerate}

\subsection{Example 1: Synthesised data}

In this example, a demonstration of the algorithm is described using some synthesised data. It must be noted that due to confidentiality agreements in effect, signals simulating the actual blade physics (and subsequent frequency estimations of the blades) cannot be reported. Instead, the edge frequency data of one blade is synthesised using a non-zero mean normal distribution. The edge frequency data of the second blade is set to be equal to that of the first blade with some additional noise. Finally, the third blade initially has the same edge frequencies of the first blade, but \lq{}deteriorates\rq{} over time by means of a negative linear gradient. That is,

\begin{equation}
f_A = \mathcal{N}(\mu_A, \sigma^2)
\end{equation}
\begin{equation}
f_B = f_A + \mathcal{N}(0, \sigma^2)
\end{equation}
\begin{equation}
f_C = f_A + d_C + \mathcal{N}(0, \sigma^2)
\end{equation}

\noindent where $d_C$ is a vector of zeros followed by a line with a negative gradient. That is,

\begin{equation} \label{eq:deterioration}
d_C = \{\boldsymbol{0}, m\boldsymbol{x} + c\}
\end{equation}

In this work, 360000 data points were generated. Only the first half of the data points of Blade C are considered \lq{}healthy\rq{}. The length of the vector of zeros in equation (\ref{eq:deterioration}) is therefore 180000. The rest of the vector $d_C$, which is indexed by the data point number, $\boldsymbol{x}$ (i.e. 180001 to 360000) has a gradient, $m$, of $-1 \times 10^{-7}$ and a y-intercept value, $c$ of 0.018. Whilst not realistic, $\mu_A$ is set to 10 and $\sigma^2$ has a value of $0.01$. 

\begin{figure}[H]
	\vspace{0pt}
	\centering
	\subfloat[]{\includegraphics[scale=0.405, trim=0cm 0cm 0cm 0cm, clip=true]{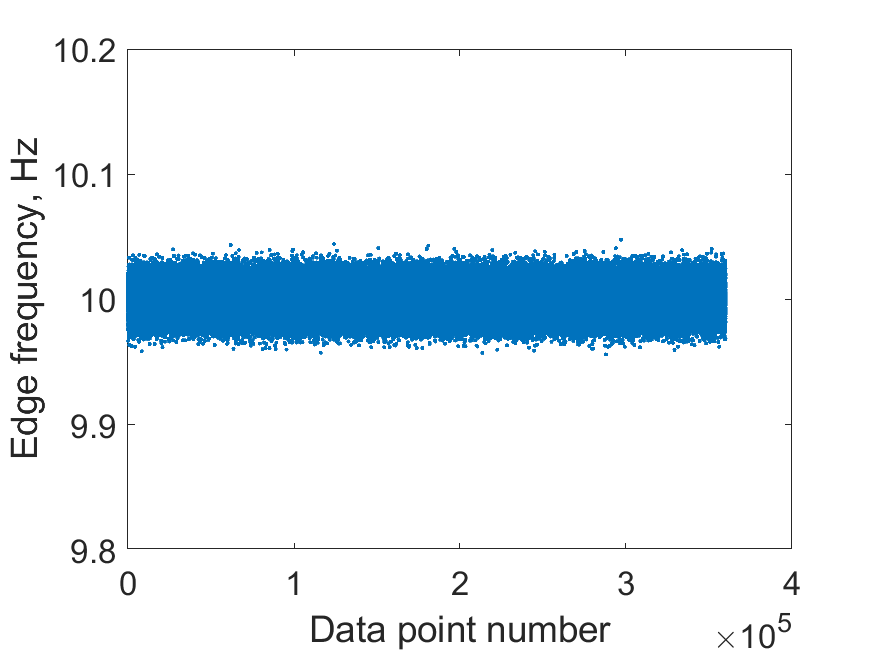}}
	\hspace{0in}
	\subfloat[]{\includegraphics[scale=0.405, trim=0cm 0cm 0cm 0cm, clip=true]{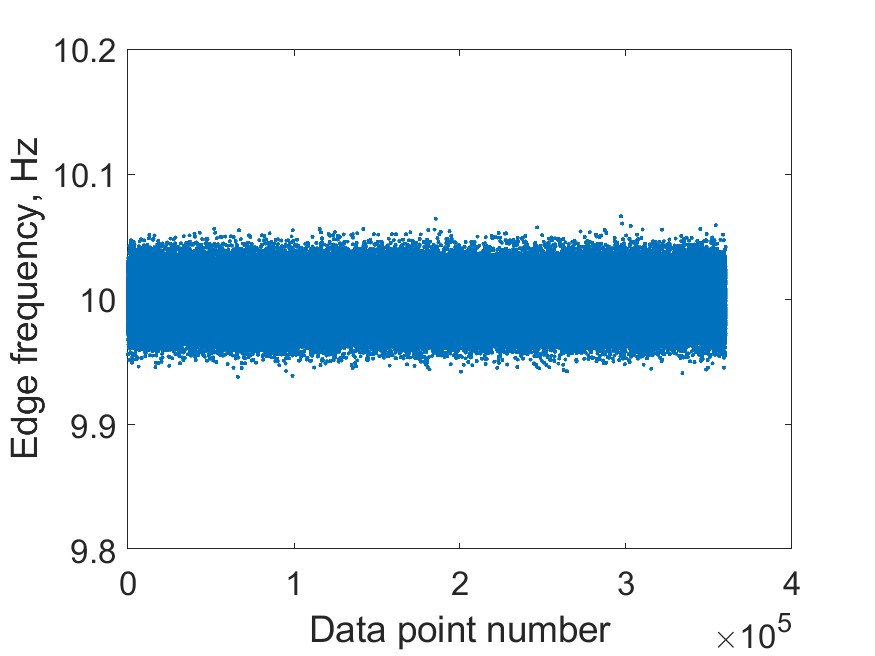}}
	\hspace{0in}
	\subfloat[]{\includegraphics[scale=0.405, trim=0cm 0cm 0cm 0cm, clip=true]{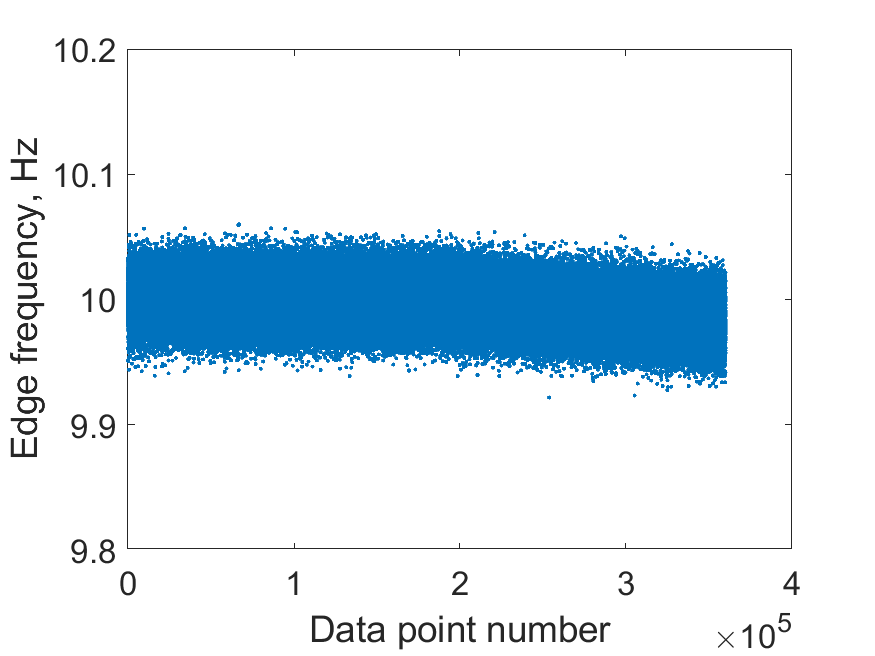}}
	\hspace{0in}
	\vspace{0in}
	\caption{Synthesised edge frequency data for: (a) Blade A, (b) Blade B, and (c) Blade C}
	\label{fig:simulated_data}
\end{figure}

Figure \ref{fig:simulated_data} illustrates the synthesised edge frequencies of the three blades.

Referring to Section \ref{sec:summary}, the above description explains the data acquisition and feature selection processes. That is, the data acquired is  As outliers do not exist on the synthesised data,t covers the first three 

\begin{figure}[H]
	\vspace{0pt}
	\centering
	\subfloat[]{\includegraphics[scale=0.405, trim=0cm 0cm 0cm 0cm, clip=true]{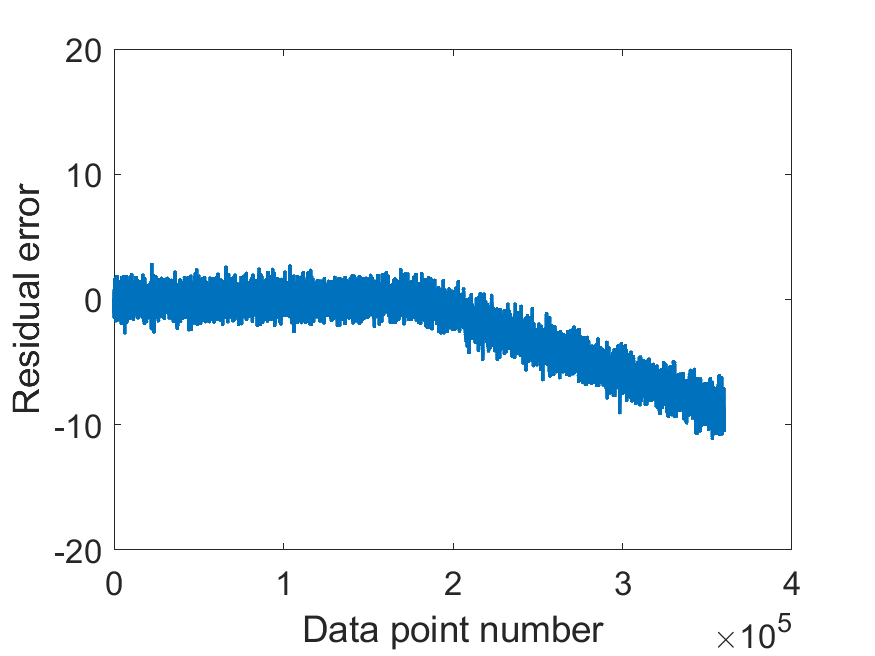}}
	\hspace{0in}
	\subfloat[]{\includegraphics[scale=0.405, trim=0cm 0cm 0cm 0cm, clip=true]{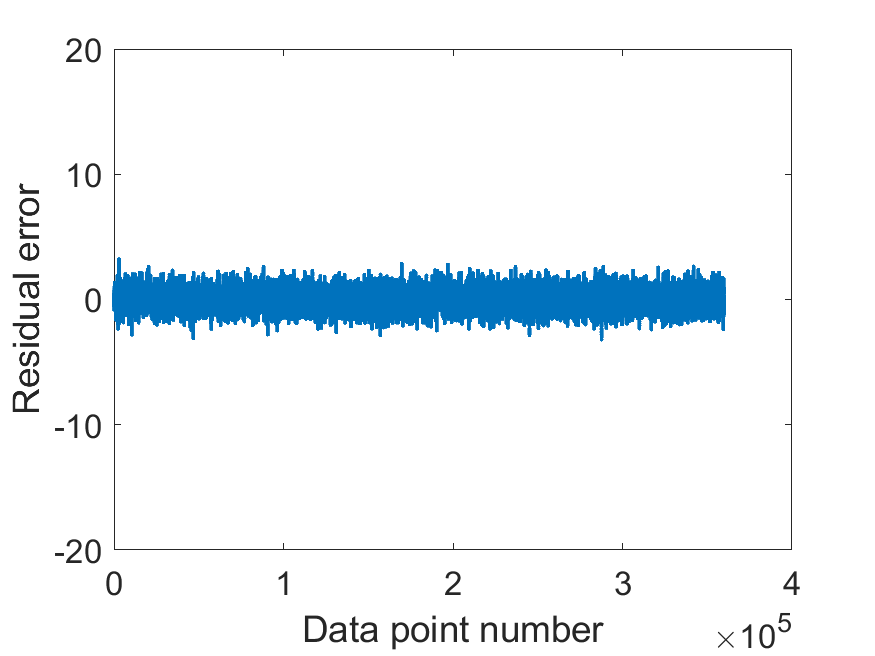}}
	\hspace{0in}
	\subfloat[]{\includegraphics[scale=0.405, trim=0cm 0cm 0cm 0cm, clip=true]{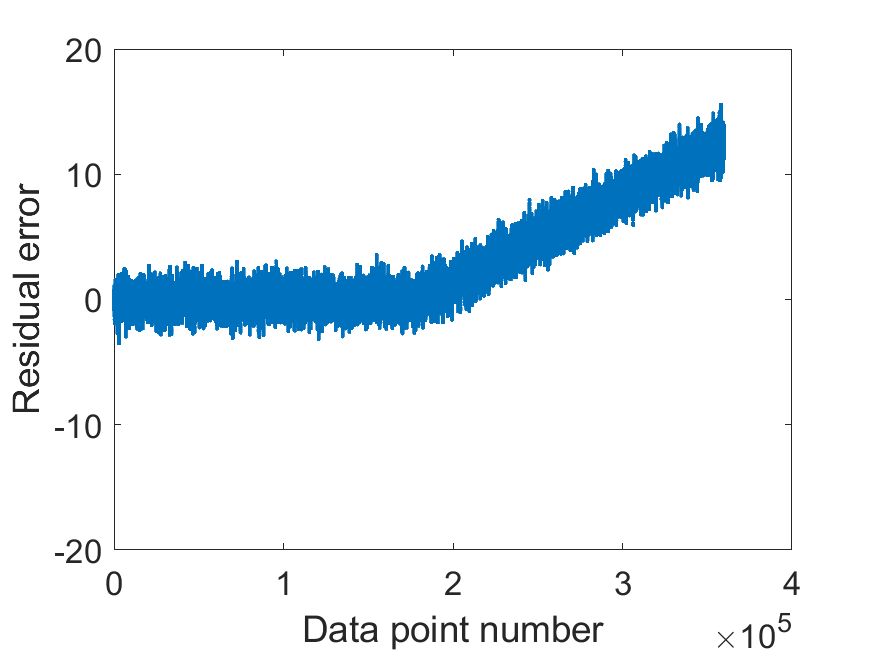}}
	\hspace{0in}
	\vspace{0in}
	\caption{Residual errors between synthesised and predicted data for: (a) Blade A, (b) Blade B, and (c) Blade C}
	\label{fig:simulated_residuals}
\end{figure}

The methodology, as described in Section \ref{sec:methodology}, is applied. Figure \ref{fig:simulated_residuals} illustrates the residual errors between the actual and predicted frequencies. Note that due to data normalisation (which is required for numerical stability, as discussed in Section \ref{sec:methodology}), the residual errors are scaled. It is clear that the \lq{}damage\rq{} is identified as the residual errors of blades A and C  diverge on the midway point. The residuals of blade B do not diverge since they do not rely on data from the damaged blade C (actual or predicted). Note that in reality, the dynamics of healthy blades may also be affected by the presence of a damaged blade, and hence it may not always be possible to conclude which blade contains damage.

\subsection{Example 2: Real turbine data}

\begin{figure}[H]
  \vspace{0pt}
  \centering
   \subfloat[]{\includegraphics[scale=0.405, trim=0cm 0cm 0cm 0cm, clip=true]{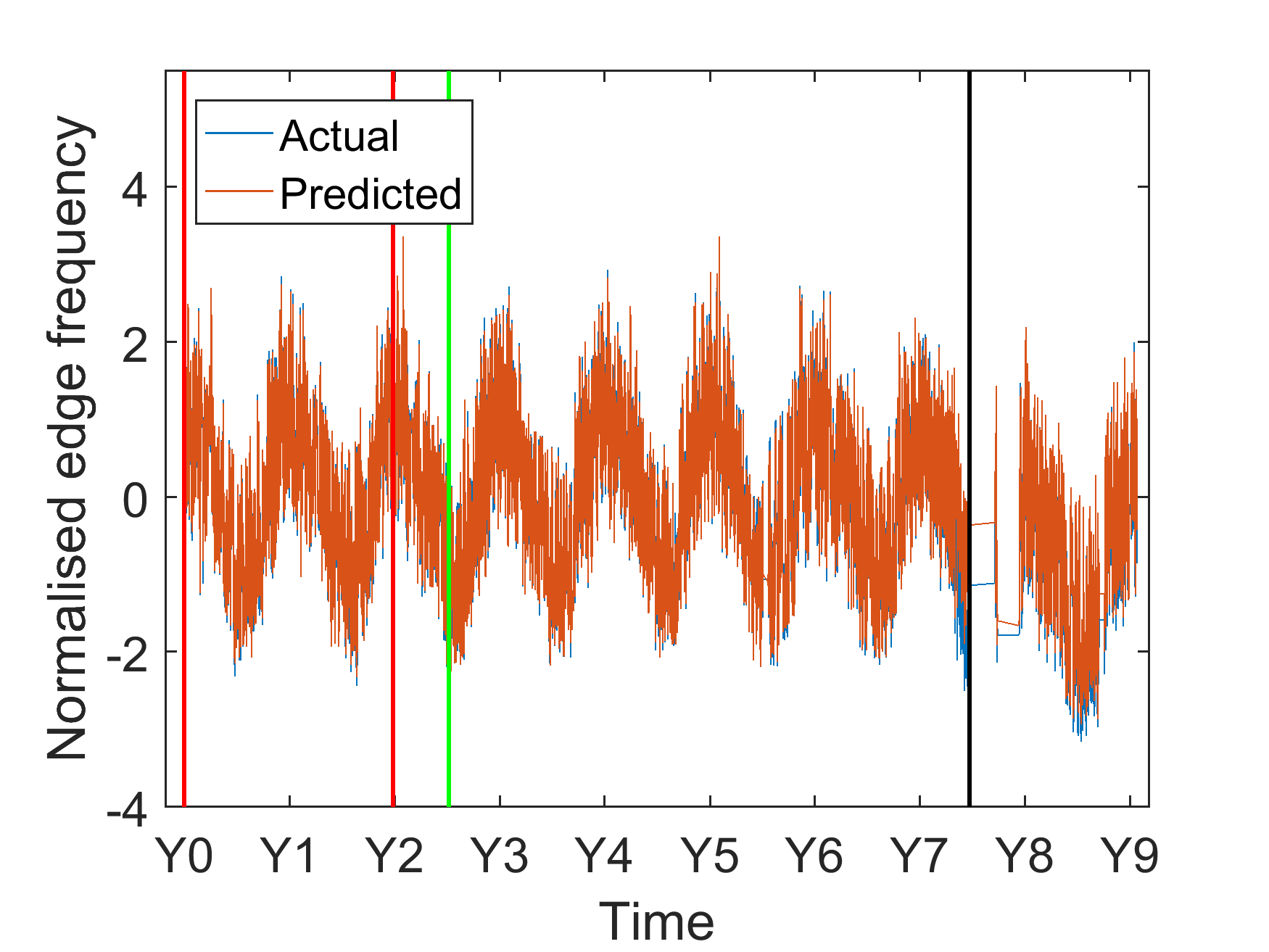}}
  \hspace{0in}
   \subfloat[]{\includegraphics[scale=0.405, trim=0cm 0cm 0cm 0cm, clip=true]{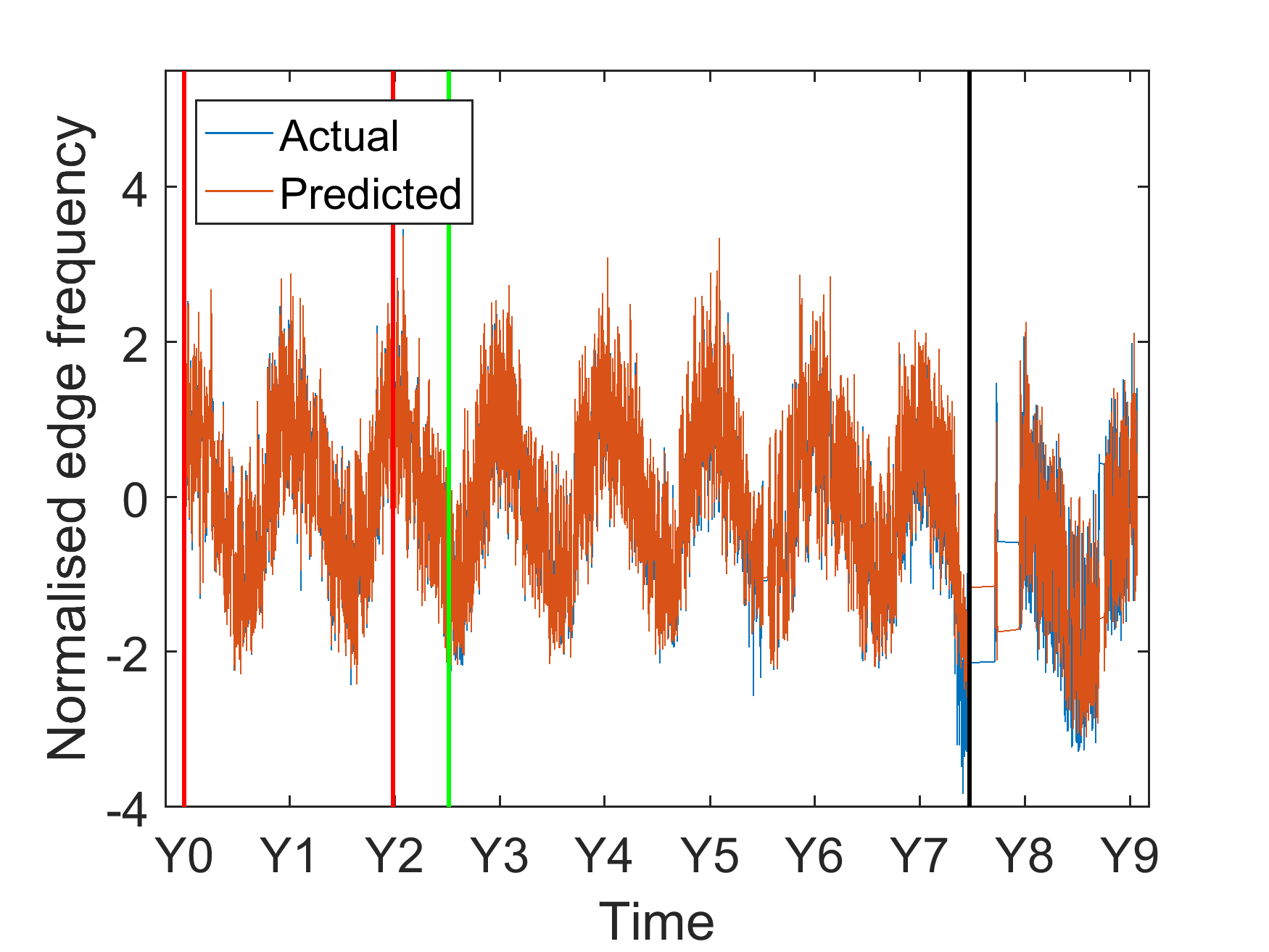}}
    \hspace{0in}
   \subfloat[]{\includegraphics[scale=0.405, trim=0cm 0cm 0cm 0cm, clip=true]{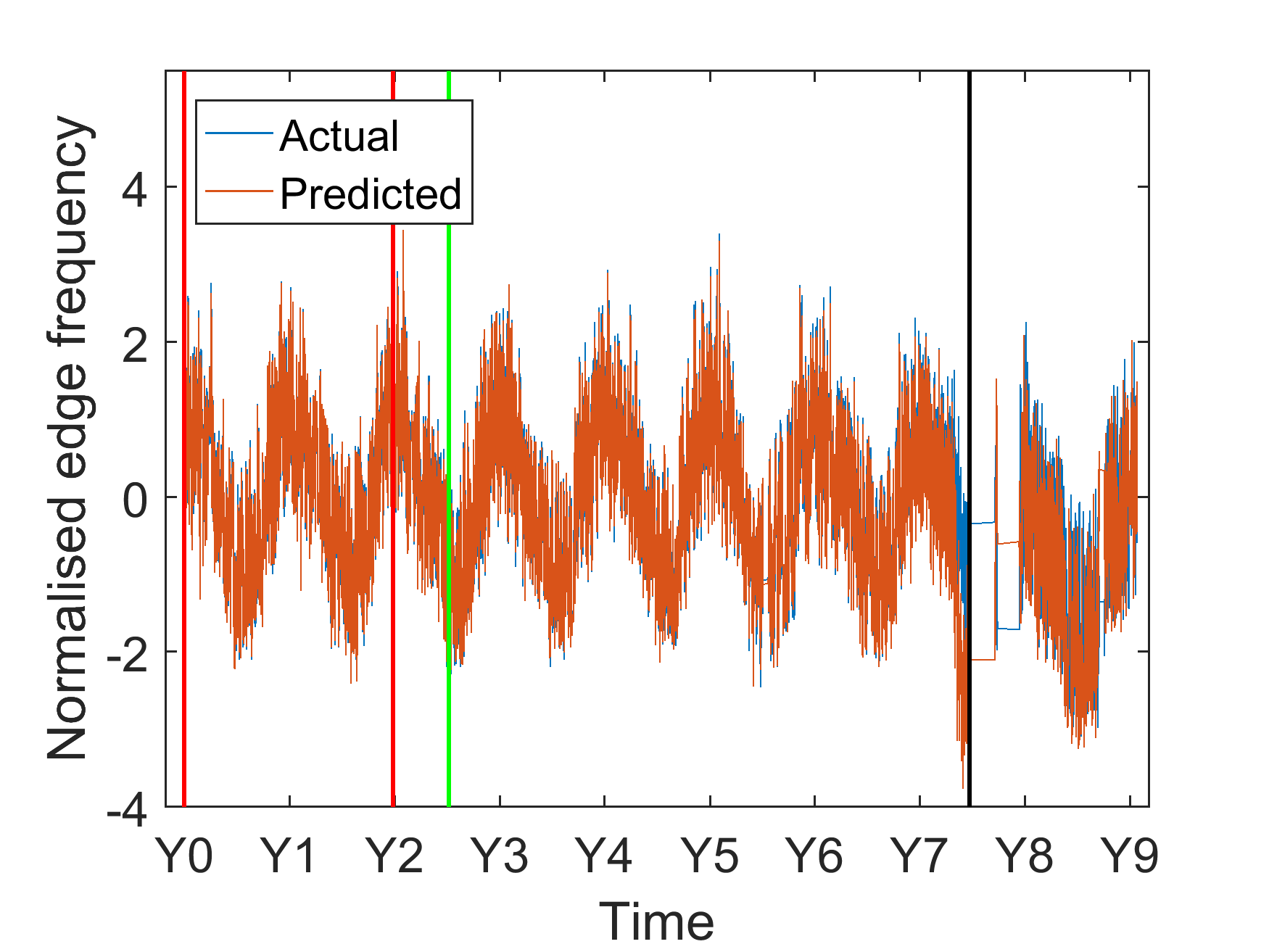}}
    \hspace{0in}
  \vspace{0in}
  \caption{Comparison between actual estimated normalised edge frequencies and GP predictions for: (a) Blade A, (b) Blade B, and (c) Blade C in Site A.}
  \label{fig:comparison_GP_actual}
\end{figure}

The correlation plots of this turbine were illustrated in Figures \ref{fig:freq_one_to_one} and \ref{fig:EOV_correlation}. In this example, one of the blades on the turbine was found to have some form of damage midway through Year 7 of turbine operation. Figure \ref{fig:comparison_GP_actual} illustrates the comparisons between the actual frequency estimations and the GP predictions for each blade. Note that on the following figures, the period between the vertical red lines illustrate the training period for the GPs, the period between the second vertical red line and the vertical green line illustrate the period the $3\sigma$ thresholds were calculated, and the vertical black lines indicate the date the damage was identified.

\begin{figure}[H]
  \vspace{0pt}
  \centering
   \subfloat[]{\includegraphics[scale=0.405, trim=0cm 0cm 0cm 0cm, clip=true]{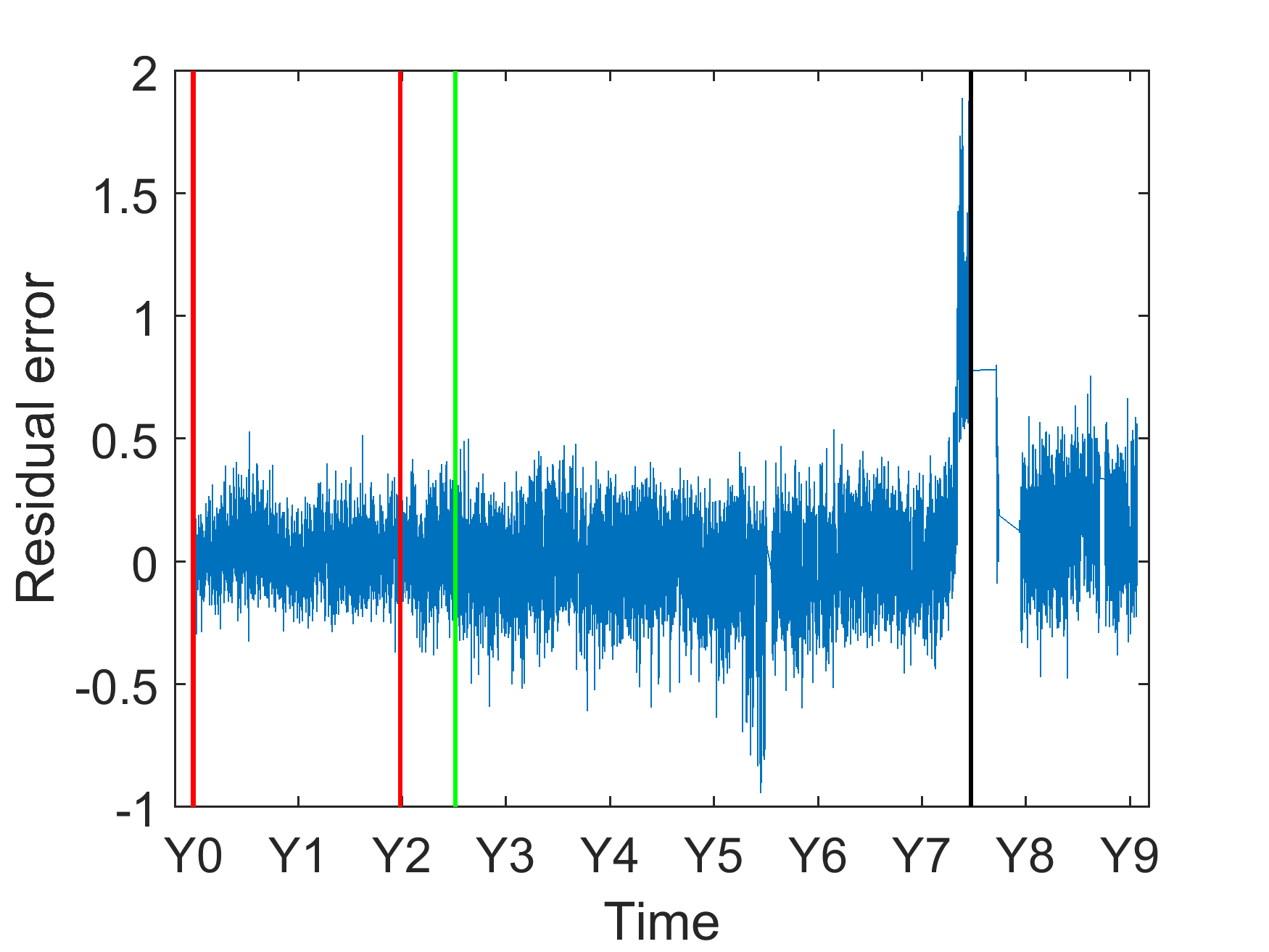}}
  \hspace{0in}
   \subfloat[]{\includegraphics[scale=0.405, trim=0cm 0cm 0cm 0cm, clip=true]{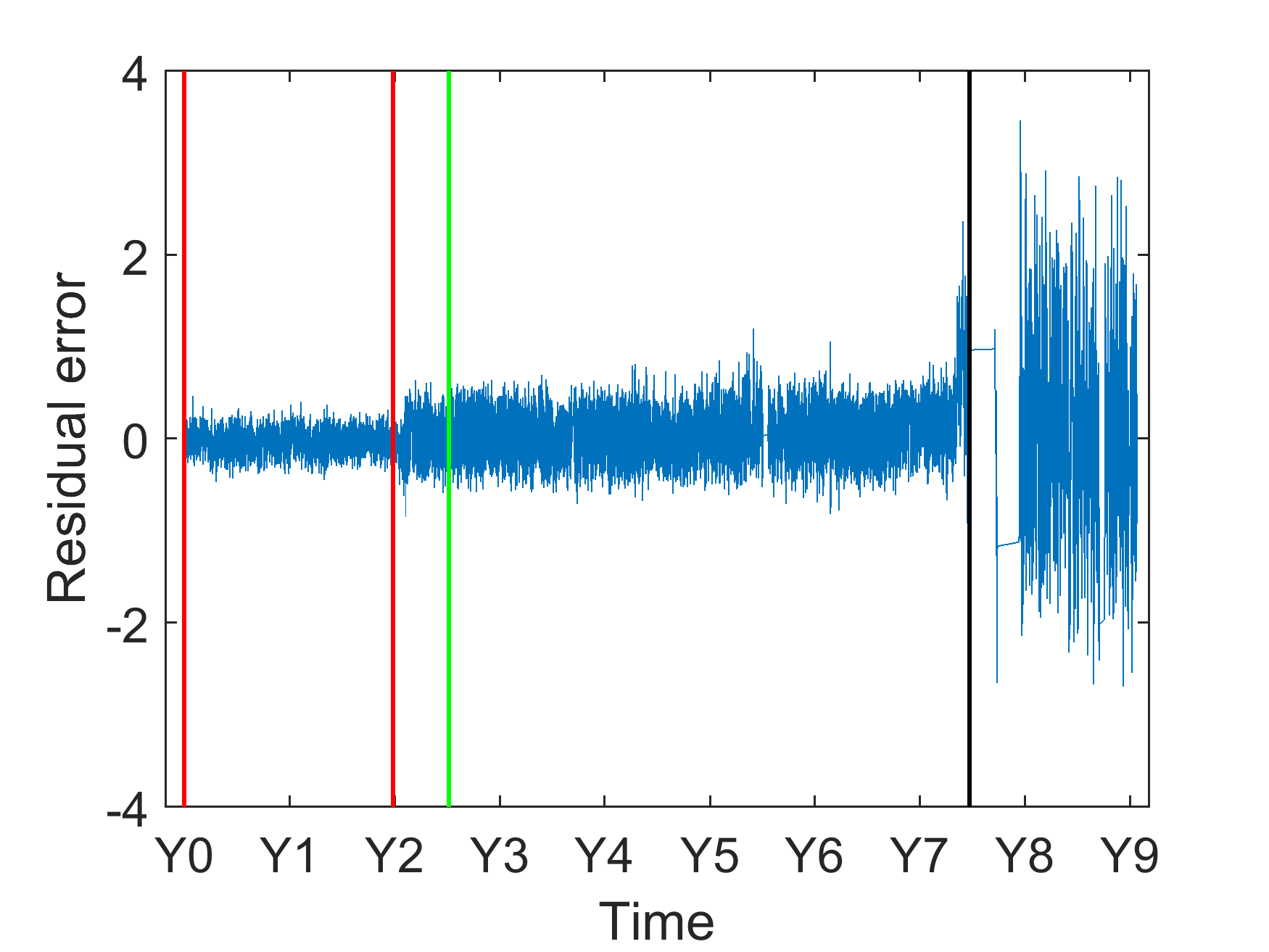}}
    \hspace{0in}
   \subfloat[]{\includegraphics[scale=0.405, trim=0cm 0cm 0cm 0cm, clip=true]{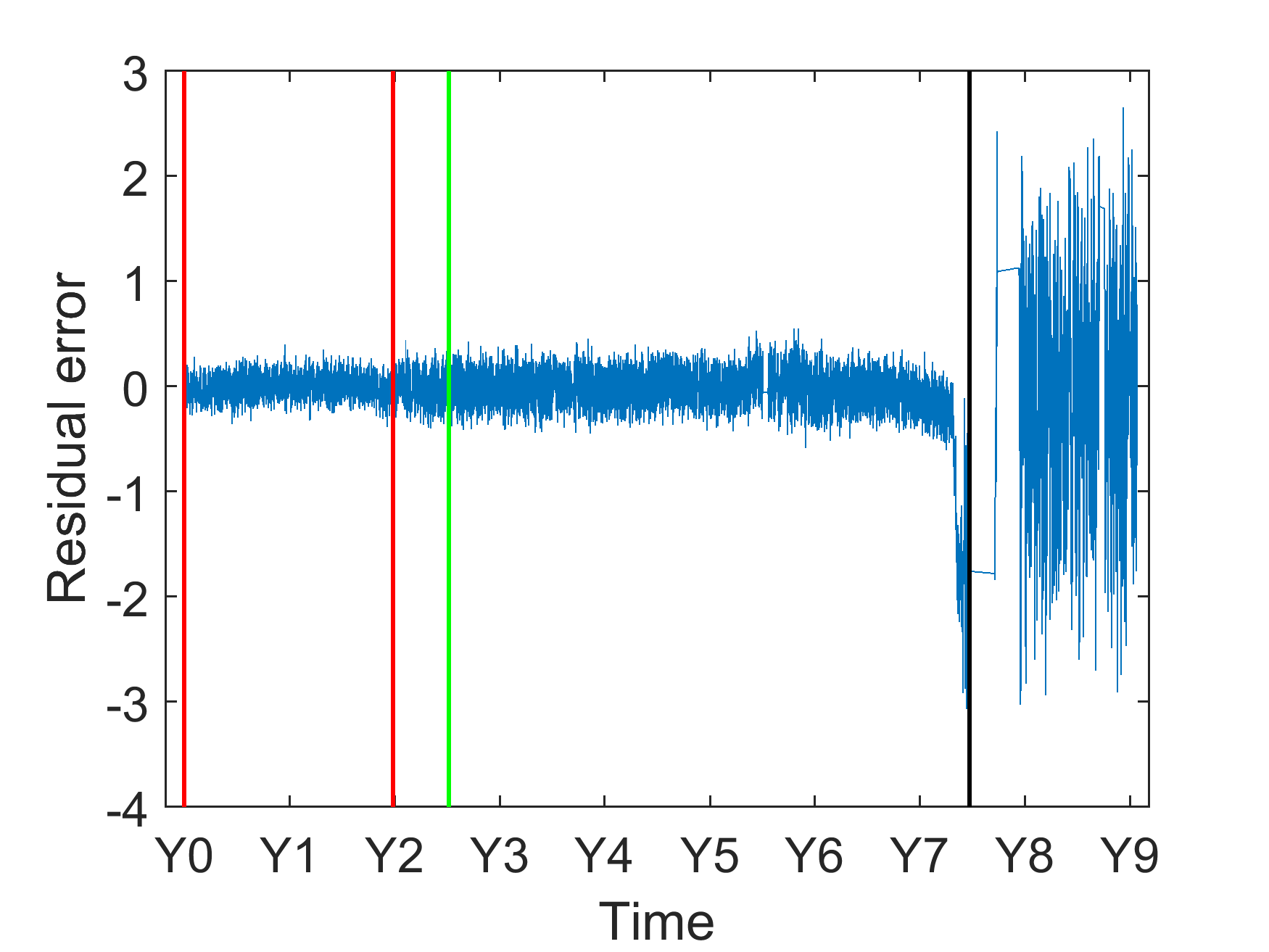}}
    \hspace{0in}
  \vspace{0in}
  \caption{Residual errors between actual estimated normalised edge frequencies and GP predictions for: (a) Blade A, (b) Blade B, and (c) Blade C in Site A.}
  \label{fig:residuals}
\end{figure}

Figures \ref{fig:residuals} (a) - (c) illustrate the residual errors corresponding to the results shown in Figures \ref{fig:comparison_GP_actual} (a) - (c), respectively.

From Figures \ref{fig:comparison_GP_actual} and \ref{fig:residuals}, it is clear that the GP predictions become less accurate as the date of damage approaches. As damage levels increase, and the properties of the blades change, the inputs to the GPs become geometrically further away than the inputs used during the training phase. Since machine learning methods in general, and GPs in particular, should not be used for extrapolation, the predictions become poorer, and hence, the residual errors grow. On closer inspection of Figure \ref{fig:residuals}, it can be seen that just prior to the sharp increases in the GP residuals, there are small mean shifts that occur.

\begin{figure}[H]
  \vspace{0pt}
  \centering
   \subfloat[]{\includegraphics[scale=0.405, trim=0cm 0cm 0cm 0cm, clip=true]{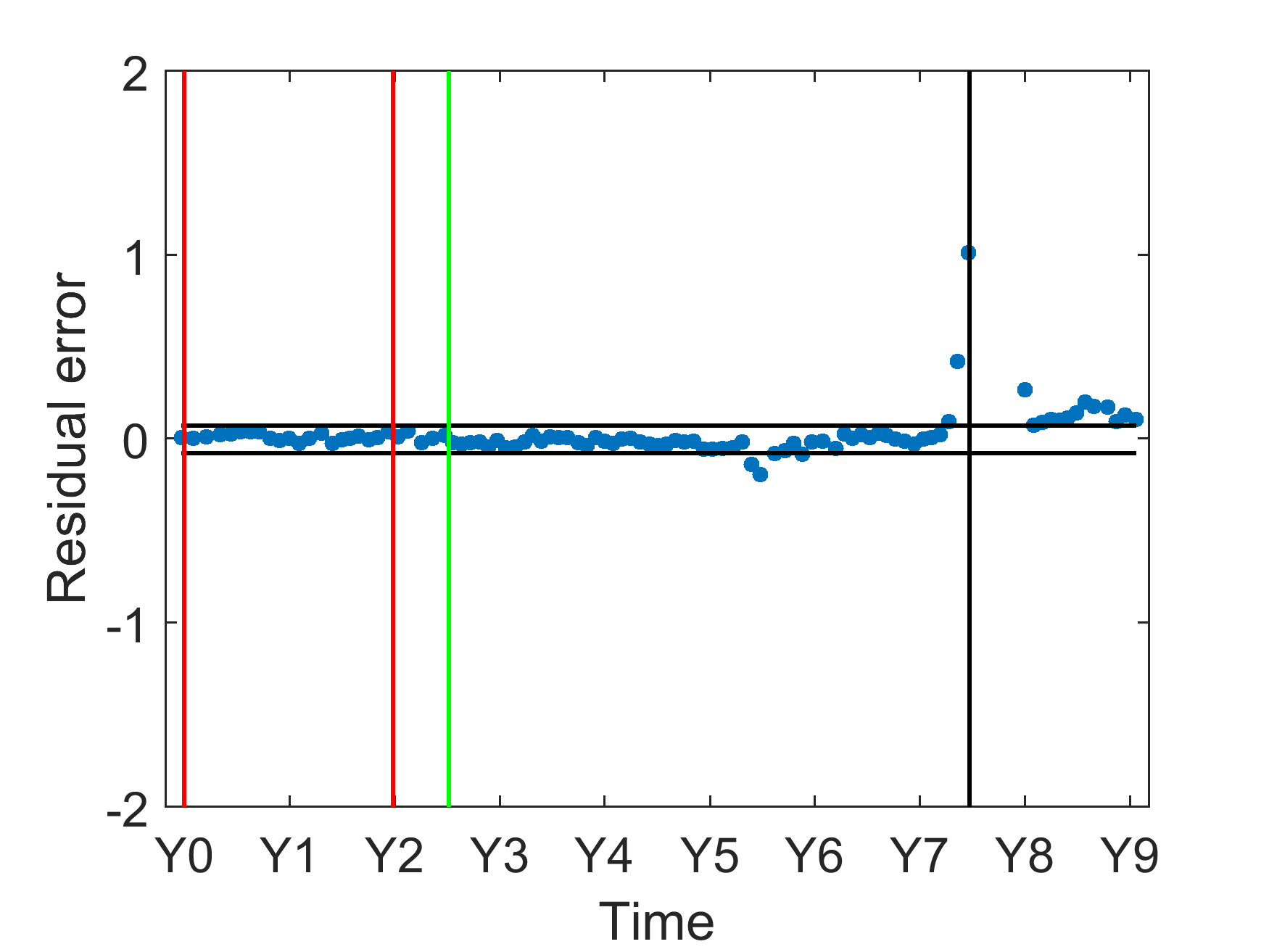}}
  \hspace{0in}
   \subfloat[]{\includegraphics[scale=0.405, trim=0cm 0cm 0cm 0cm, clip=true]{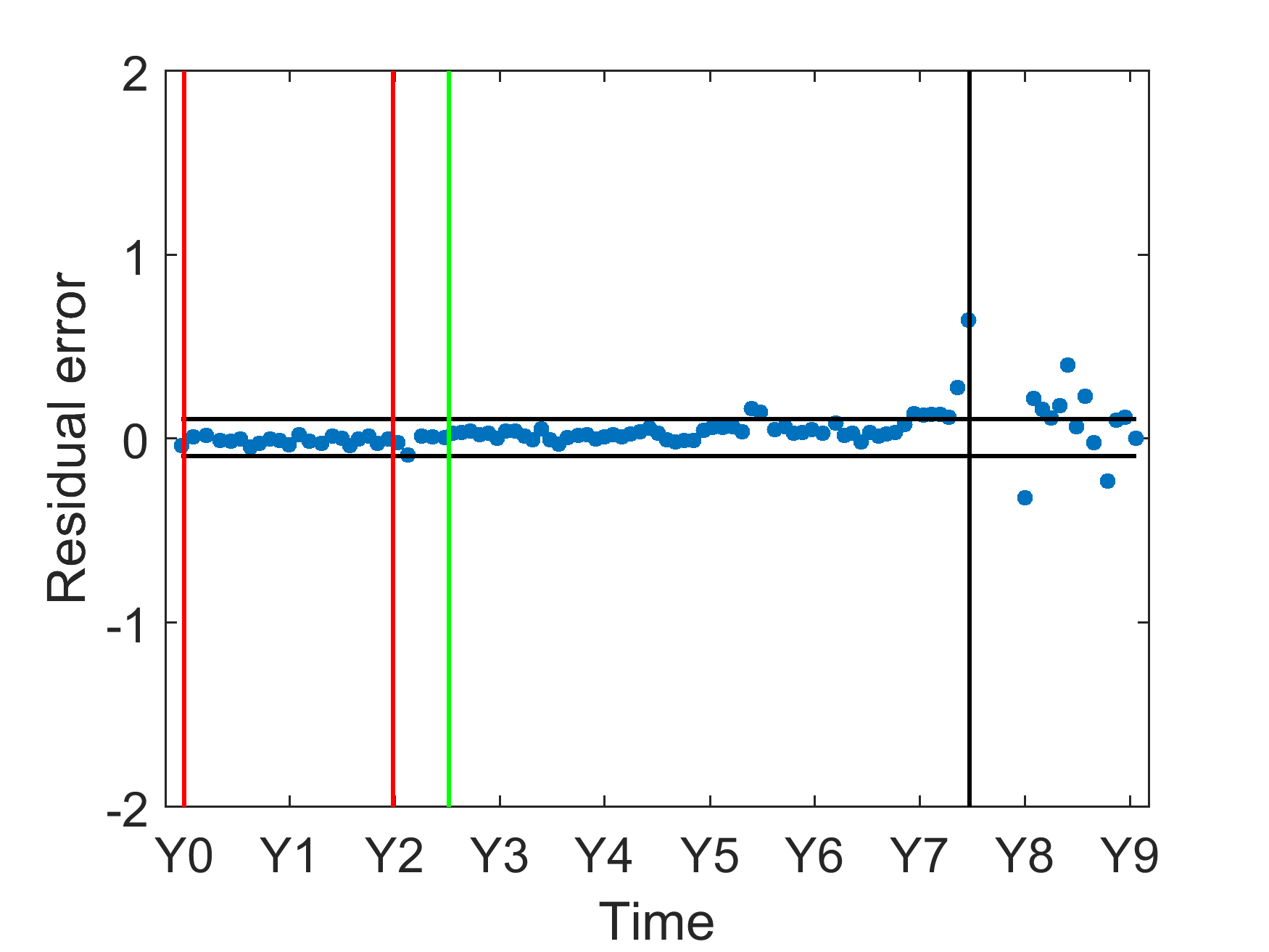}}
    \hspace{0in}
   \subfloat[]{\includegraphics[scale=0.405, trim=0cm 0cm 0cm 0cm, clip=true]{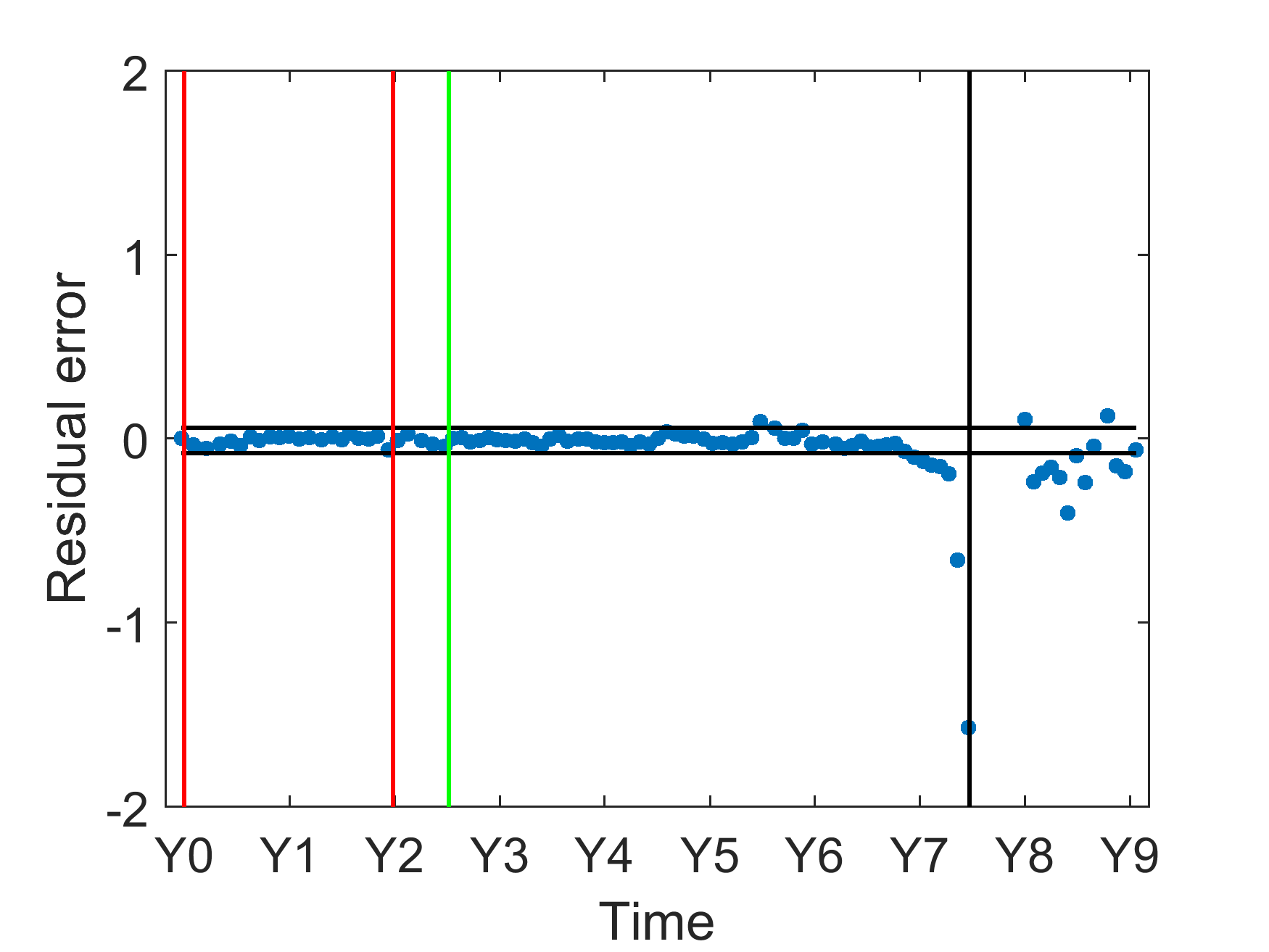}}
    \hspace{0in}
  \vspace{0in}
  \caption{X-bar control charts of GP residuals for: (a) Blade A, (b) Blade B, and (c) Blade C in Site A.}
  \label{fig:x_bar_siteA}
\end{figure}

The corresponding X-bar control charts for Site A are shown in Figure \ref{fig:x_bar_siteA}. The averages were completed once every four weeks.

The horizontal black lines indicate the calculated thresholds of the X-bar control charts. On analysing these charts in Figure \ref{fig:x_bar_siteA}, it can be seen that the GP residuals begin to indicate that the blade properties have changed roughly 6 months before the blade was remedied/replaced.

It is noteworthy that following the identification of damage, and the corresponding remedial action, there is a break down in the correlation between the pairs of blades. This is because the mechanical properties of the repaired/replaced blade are now different. Hence, whilst the residuals approach the original thresholds, they do not settle within them. This fact indicates that every time a remedial activity takes place, retraining is essential.

\subsection{Example 3: Real turbine data}

\begin{figure}[H]
	\vspace{0pt}
	\centering
	\subfloat[]{\includegraphics[scale=0.405, trim=0cm 0cm 0cm 0cm, clip=true]{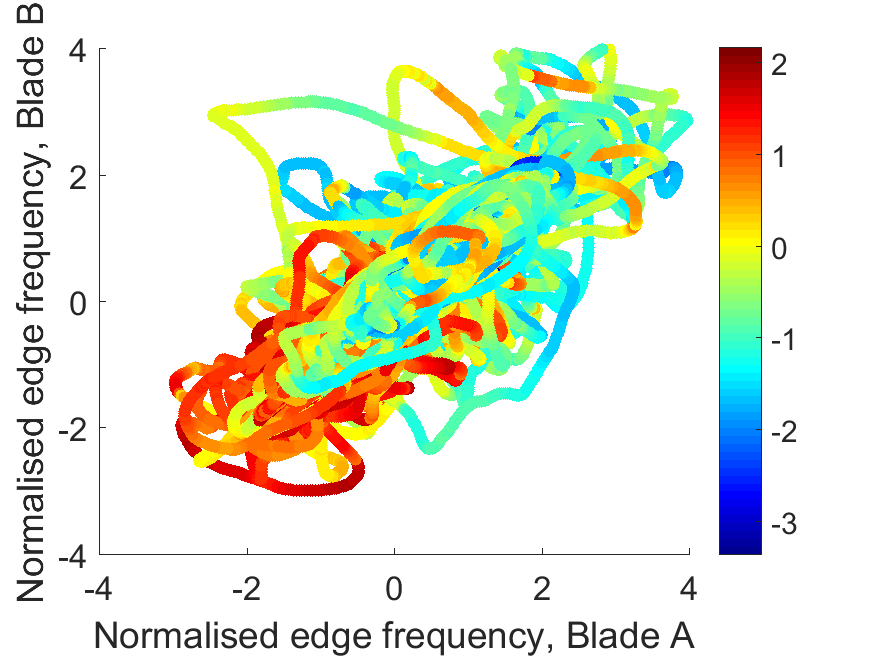}}
	\hspace{0in}
	\subfloat[]{\includegraphics[scale=0.405, trim=0cm 0cm 0cm 0cm, clip=true]{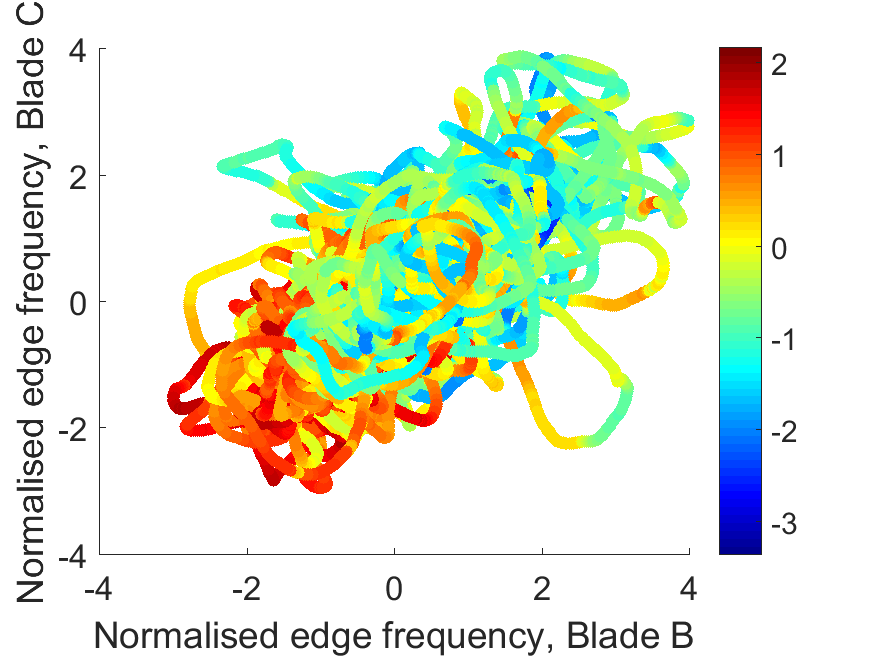}}
	\hspace{0in}
	\subfloat[]{\includegraphics[scale=0.405, trim=0cm 0cm 0cm 0cm, clip=true]{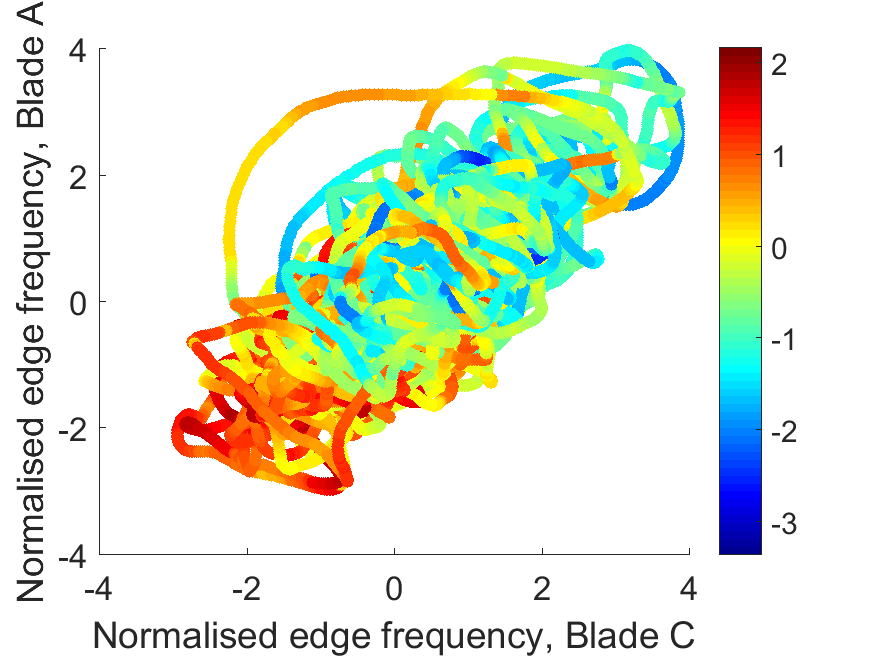}}
	\hspace{0in}
	\vspace{0in}
	\caption{Correlation between normalised edge frequencies for: (a) Blade A vs. Blade B, (b) Blade B vs. Blade C, and (c) Blade C vs. Blade A in Site B.}
	\label{fig:freq_one_to_one_temp_siteB}
\end{figure}

The correlation plots for this turbine are shown in Figure \ref{fig:freq_one_to_one_temp_siteB}. Note that the temperature scales shown in these plots are normalised. In this example, all three correlation plots are noisy, although there is the expected positive correlation. The temperature effects are noticeable - at lower temperatures, when the blades are stiffer, the edge frequencies are higher, whereas at higher temperatures, the opposite is true.

Figure \ref{fig:freq_one_to_one_temp_siteB} illustrates how the GP predictions vary with the input edge frequencies and temperature. It is interesting to note how the GPs have identified a hidden structure within the noisy data - there are various bands of predictions across the range of temperatures. These bands are both linear (captured by the linear kernel) and nonlinear (captured by the squared-exponential kernel).

\begin{figure}[H]
	\vspace{0pt}
	\centering
	\subfloat[]{\includegraphics[scale=0.405, trim=0cm 0cm 0cm 0cm, clip=true]{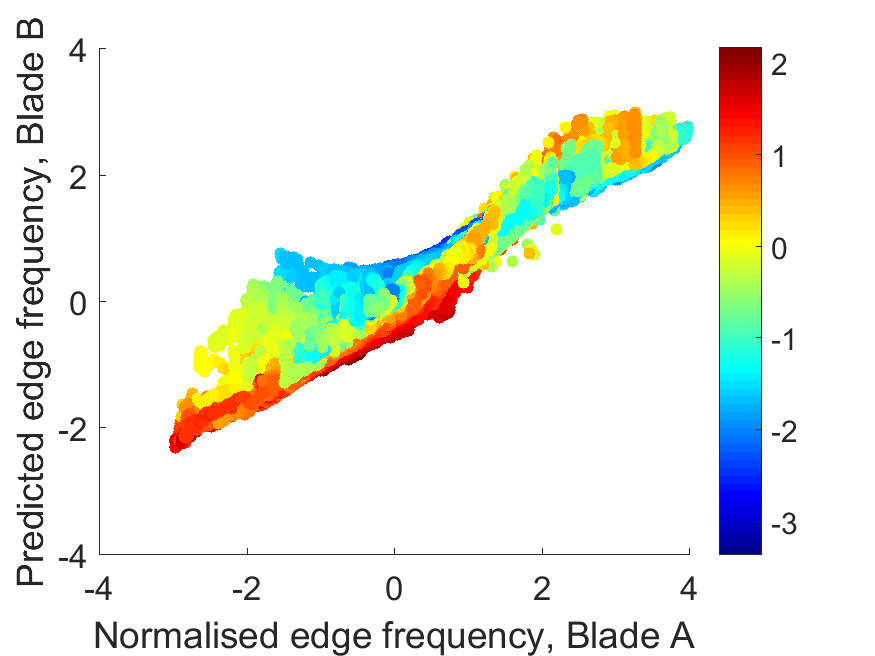}}
	\hspace{0in}
	\subfloat[]{\includegraphics[scale=0.405, trim=0cm 0cm 0cm 0cm, clip=true]{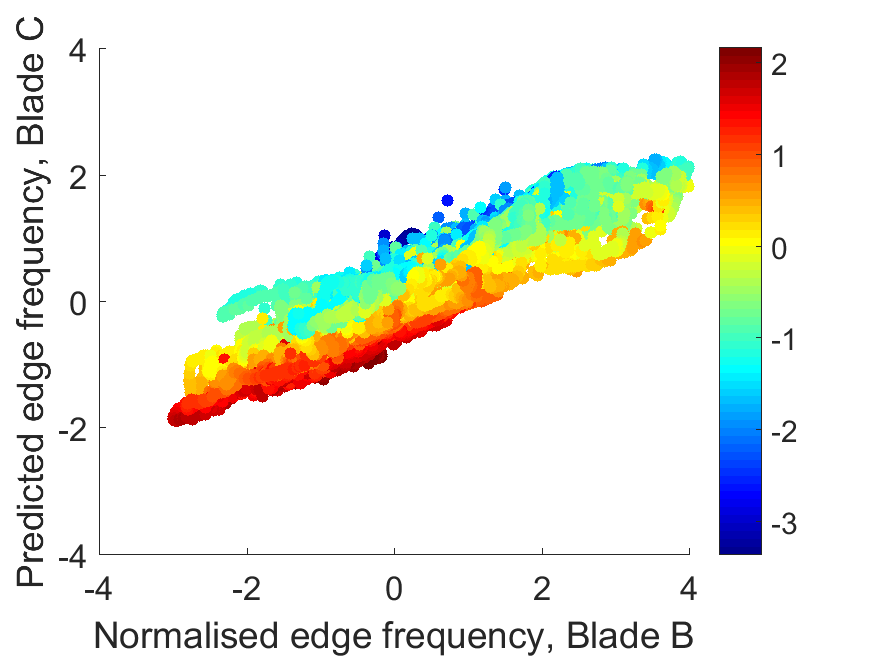}}
	\hspace{0in}
	\subfloat[]{\includegraphics[scale=0.405, trim=0cm 0cm 0cm 0cm, clip=true]{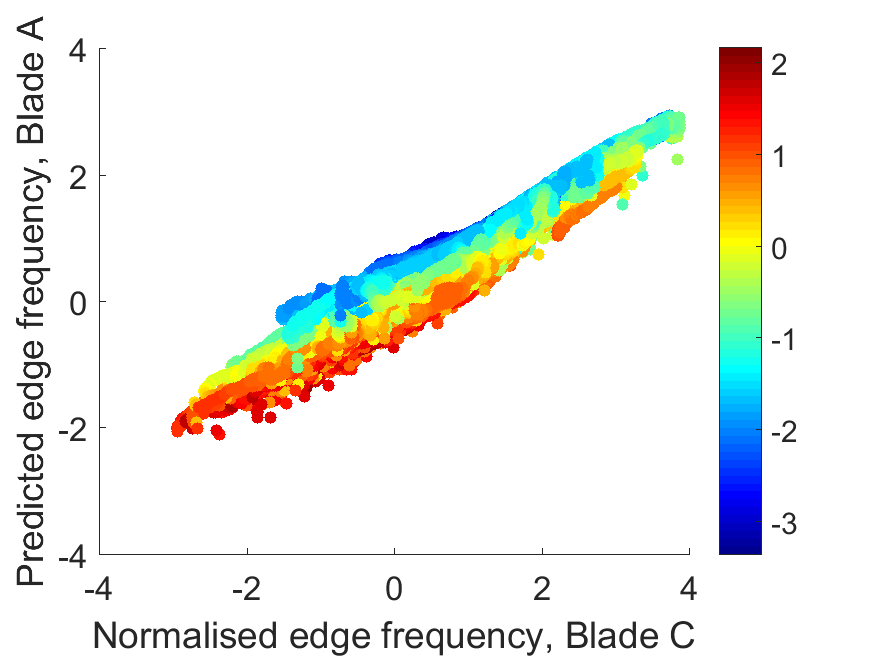}}
	\hspace{0in}
	\vspace{0in}
	\caption{Correlation between normalised edge frequencies for: (a) Blade A vs. Blade B, (b) Blade B vs. Blade C, and (c) Blade C vs. Blade A in Site B.}
	\label{fig:freq_one_to_one_temp_GP_siteB}
\end{figure}

\begin{figure}[H]
	\vspace{0pt}
	\centering
	\subfloat[]{\includegraphics[scale=0.405, trim=0cm 0cm 0cm 0cm, clip=true]{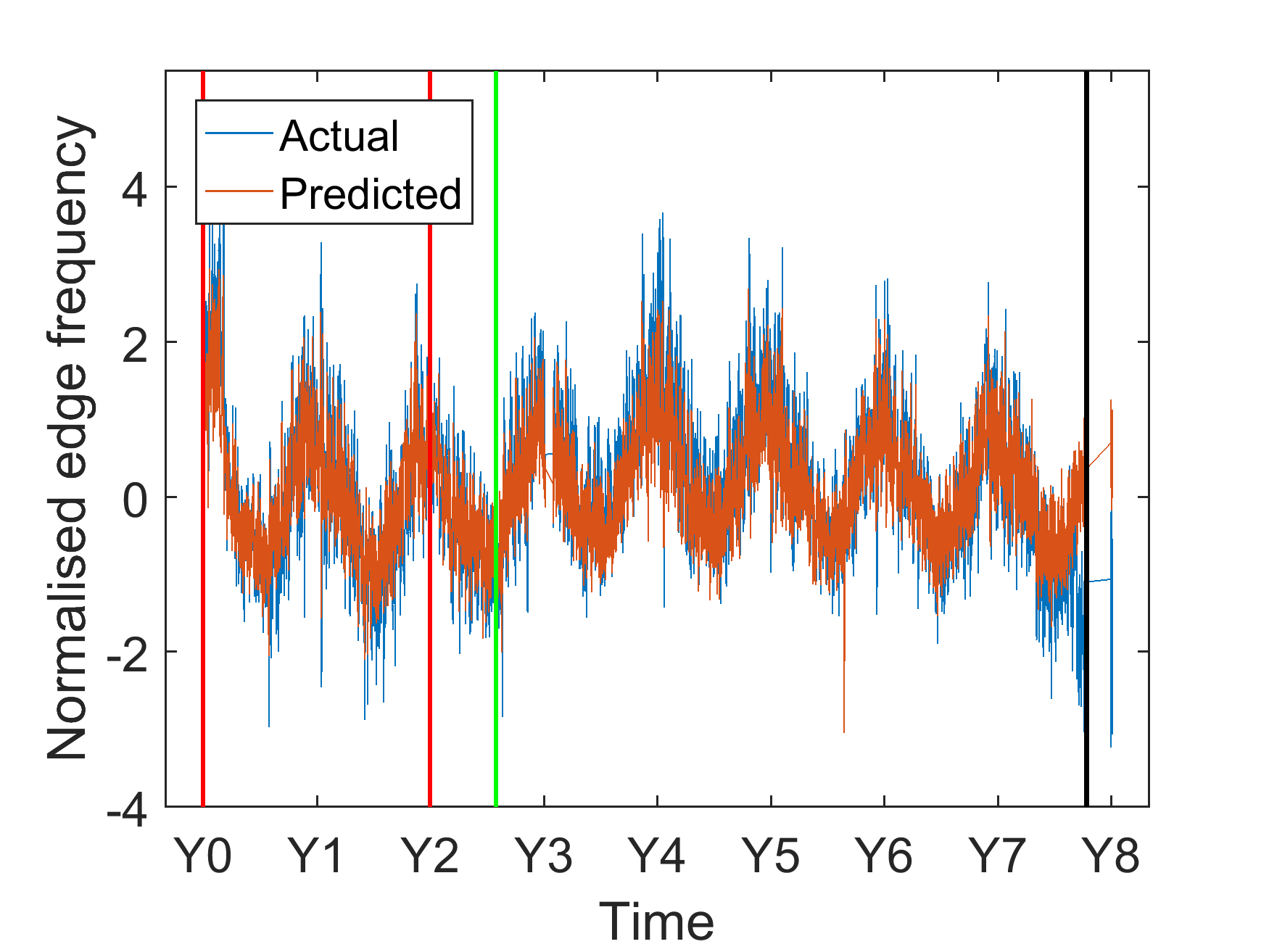}}
	\hspace{0in}
	\subfloat[]{\includegraphics[scale=0.405, trim=0cm 0cm 0cm 0cm, clip=true]{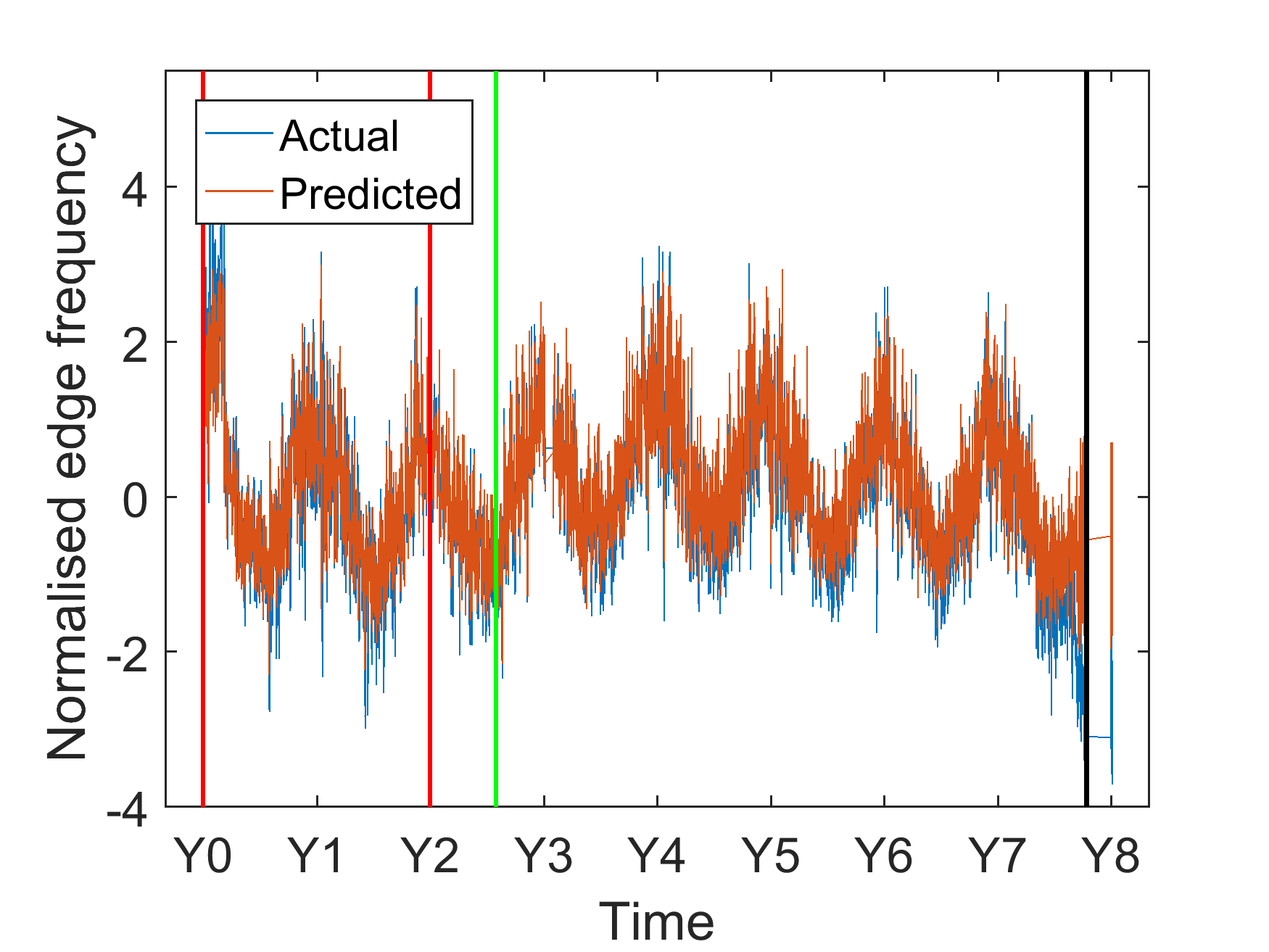}}
	\hspace{0in}
	\subfloat[]{\includegraphics[scale=0.405, trim=0cm 0cm 0cm 0cm, clip=true]{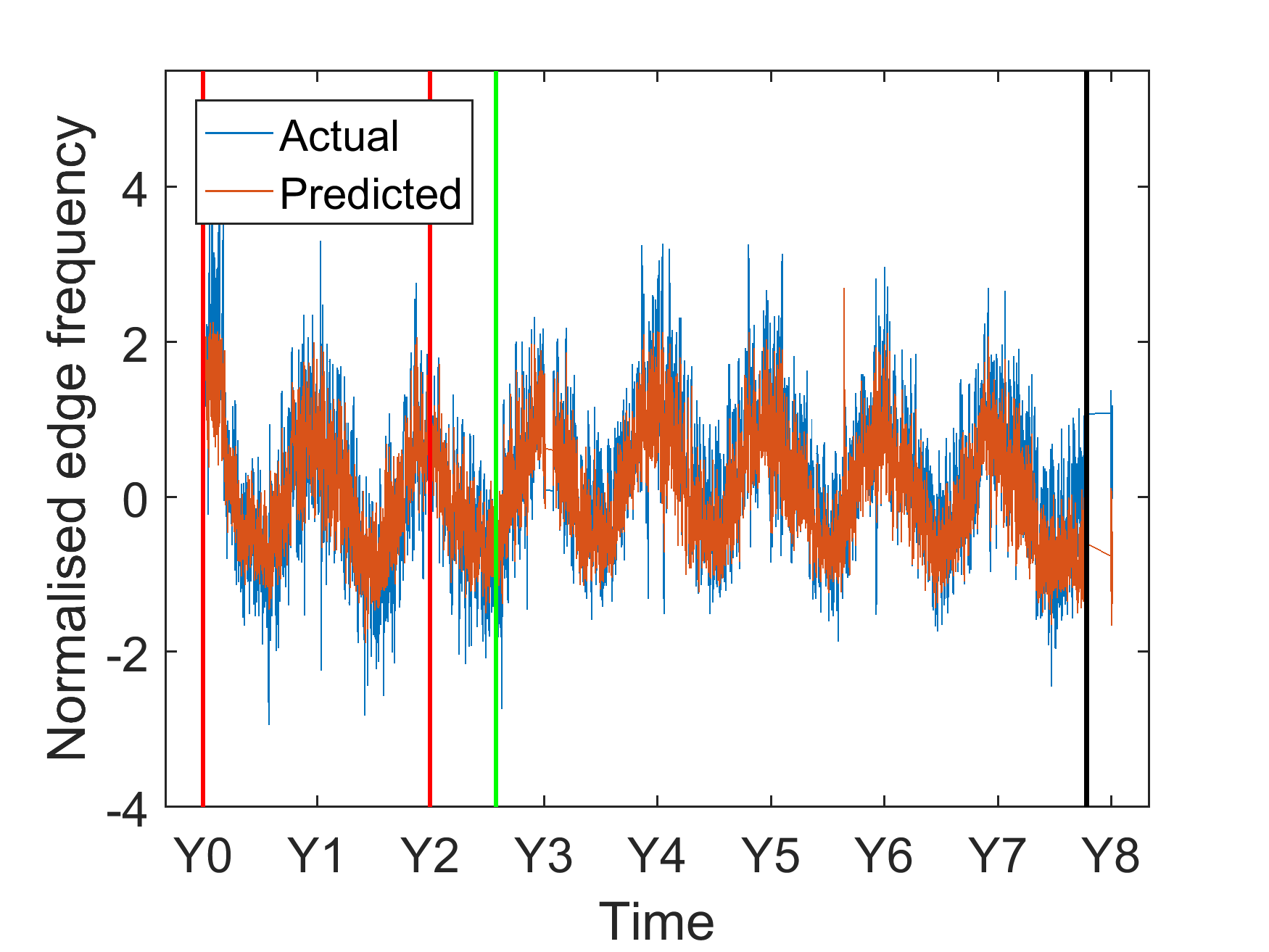}}
	\hspace{0in}
	\vspace{0in}
	\caption{Comparison between actual estimated normalised edge frequencies, and GP predictions for: (a) Blade A, (b) Blade B, and (c) Blade C in Site B.}
	\label{fig:comparison_GP_actual_siteB}
\end{figure}

\begin{figure}[H]
	\vspace{0pt}
	\centering
	\subfloat[]{\includegraphics[scale=0.405, trim=0cm 0cm 0cm 0cm, clip=true]{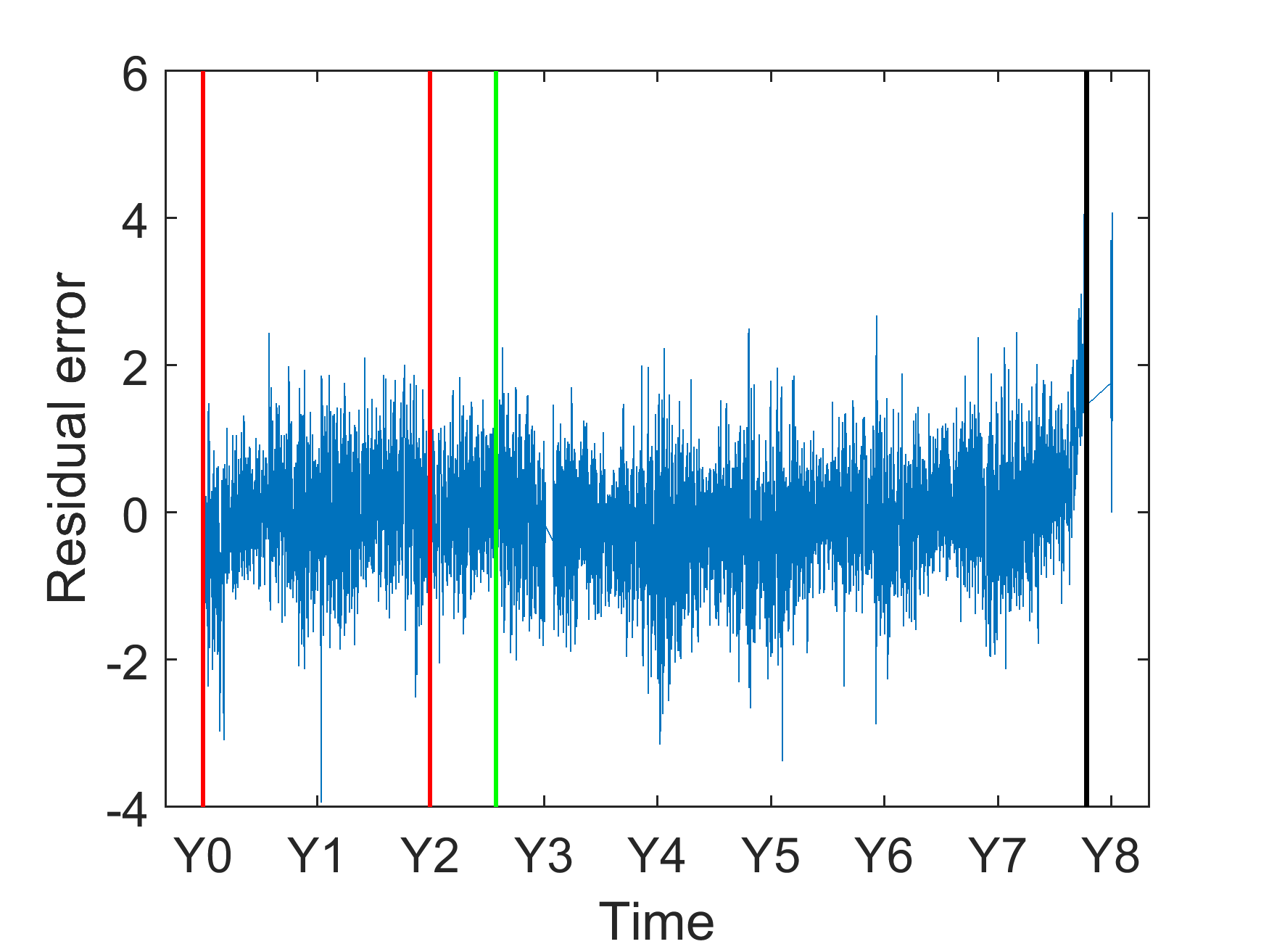}}
	\hspace{0in}
	\subfloat[]{\includegraphics[scale=0.405, trim=0cm 0cm 0cm 0cm, clip=true]{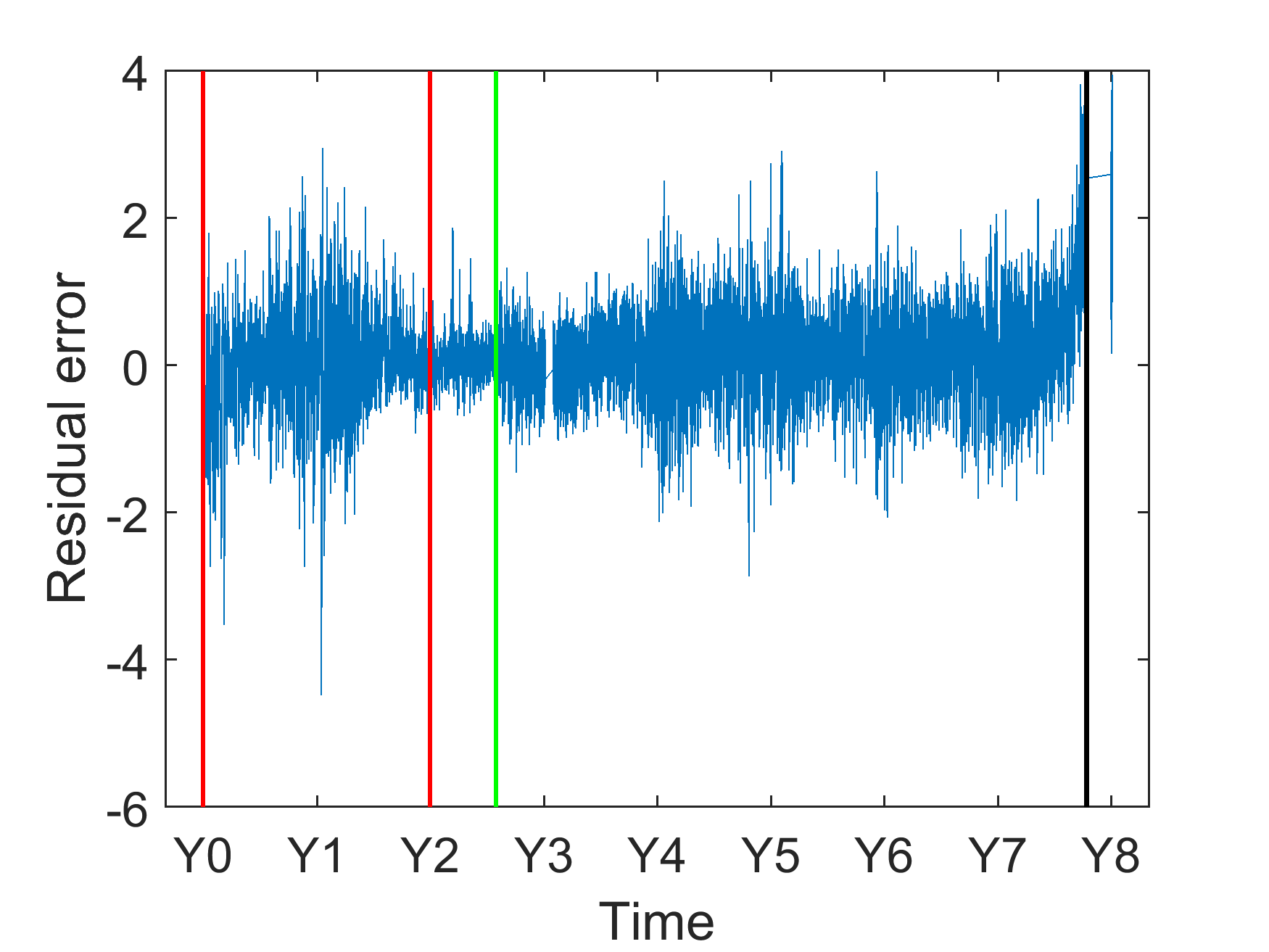}}
	\hspace{0in}
	\subfloat[]{\includegraphics[scale=0.405, trim=0cm 0cm 0cm 0cm, clip=true]{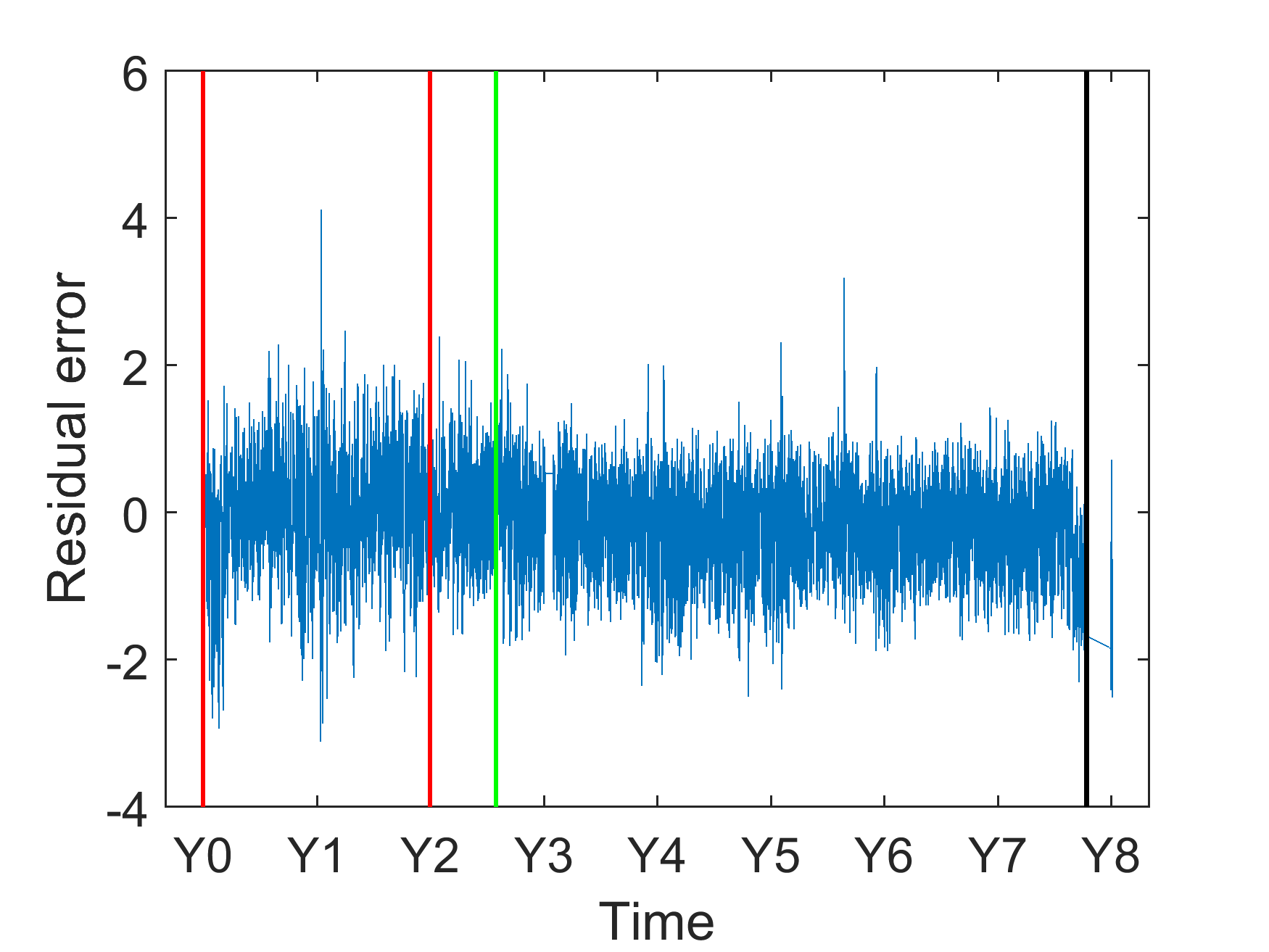}}
	\hspace{0in}
	\vspace{0in}
	\caption{Residual errors between actual estimated normalised edge frequencies, and GP predictions for: (a) Blade A, (b) Blade B, and (c) Blade C in Site B.}
	\label{fig:residuals_siteB}
\end{figure}

Figures \ref{fig:comparison_GP_actual_siteB} and \ref{fig:residuals_siteB} compare the estimated and predicted edge frequencies. Once again, the seasonal variations are captured well. However, the amplitudes of the residual errors are high compared to those seen in Example 2, which is expected given the noisy estimates and correlations. The residuals in Figure \ref{fig:residuals_siteB} (a) and (b) indicate that an event occurred that changed the structural properties, and correlations between the blades around Year 4 as shown by the change in structure of the residuals. As the damage event approaches, there is a monotonic change in the residuals in Figure \ref{fig:residuals_siteB}.

\begin{figure}[H]
  \vspace{0pt}
  \centering
   \subfloat[]{\includegraphics[scale=0.405, trim=0cm 0cm 0cm 0cm, clip=true]{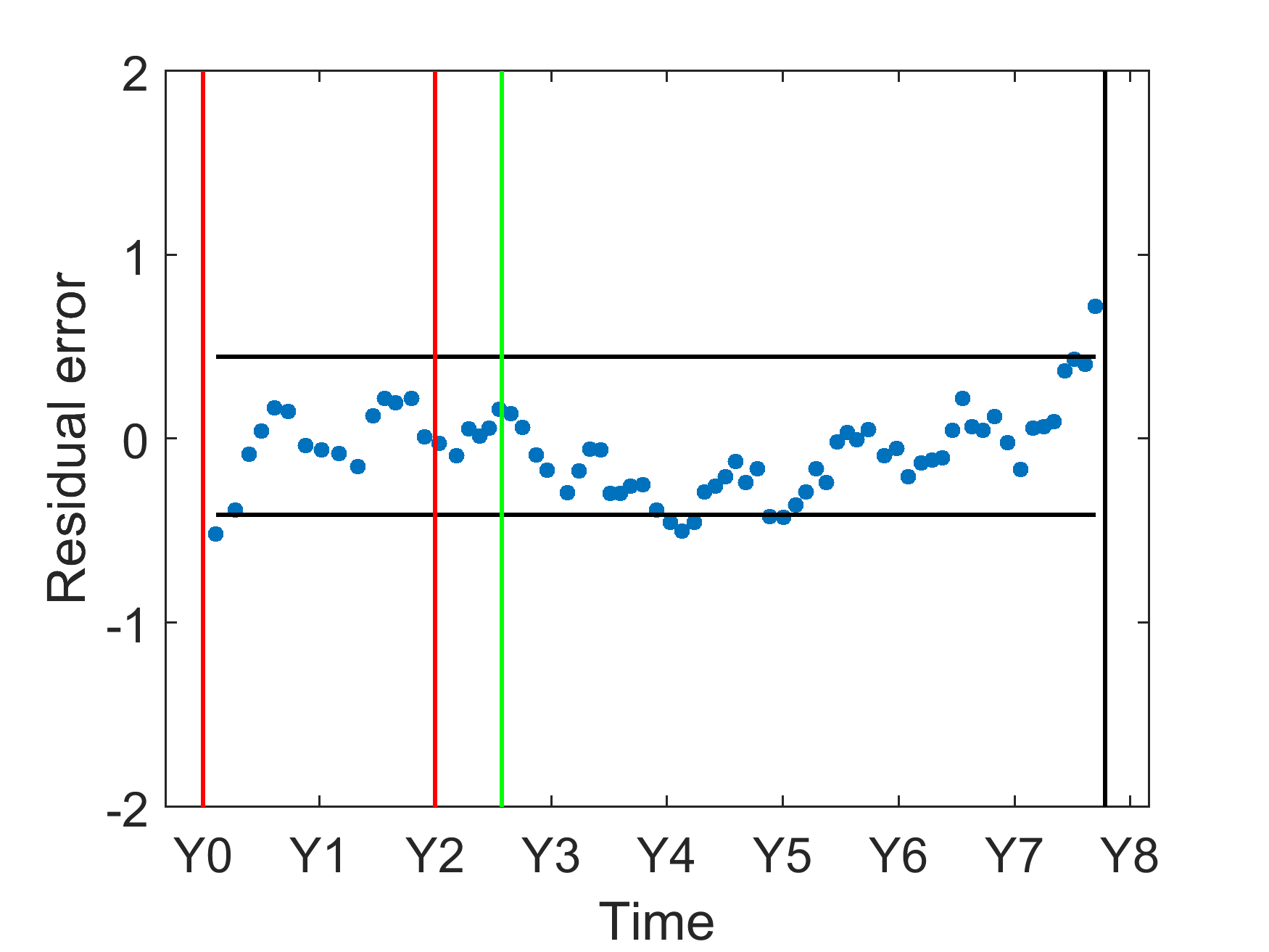}}
  \hspace{0in}
   \subfloat[]{\includegraphics[scale=0.405, trim=0cm 0cm 0cm 0cm, clip=true]{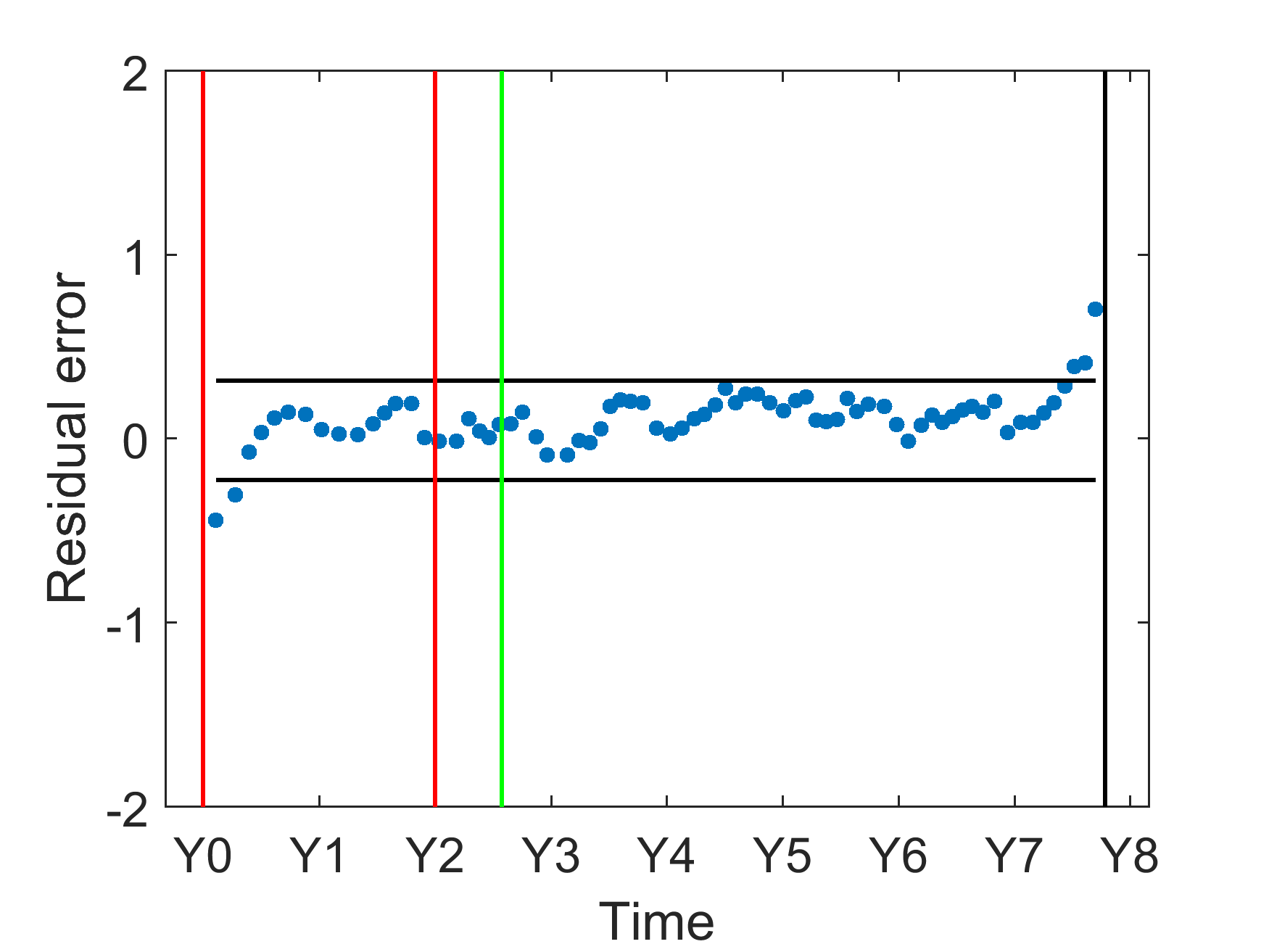}}
    \hspace{0in}
   \subfloat[]{\includegraphics[scale=0.405, trim=0cm 0cm 0cm 0cm, clip=true]{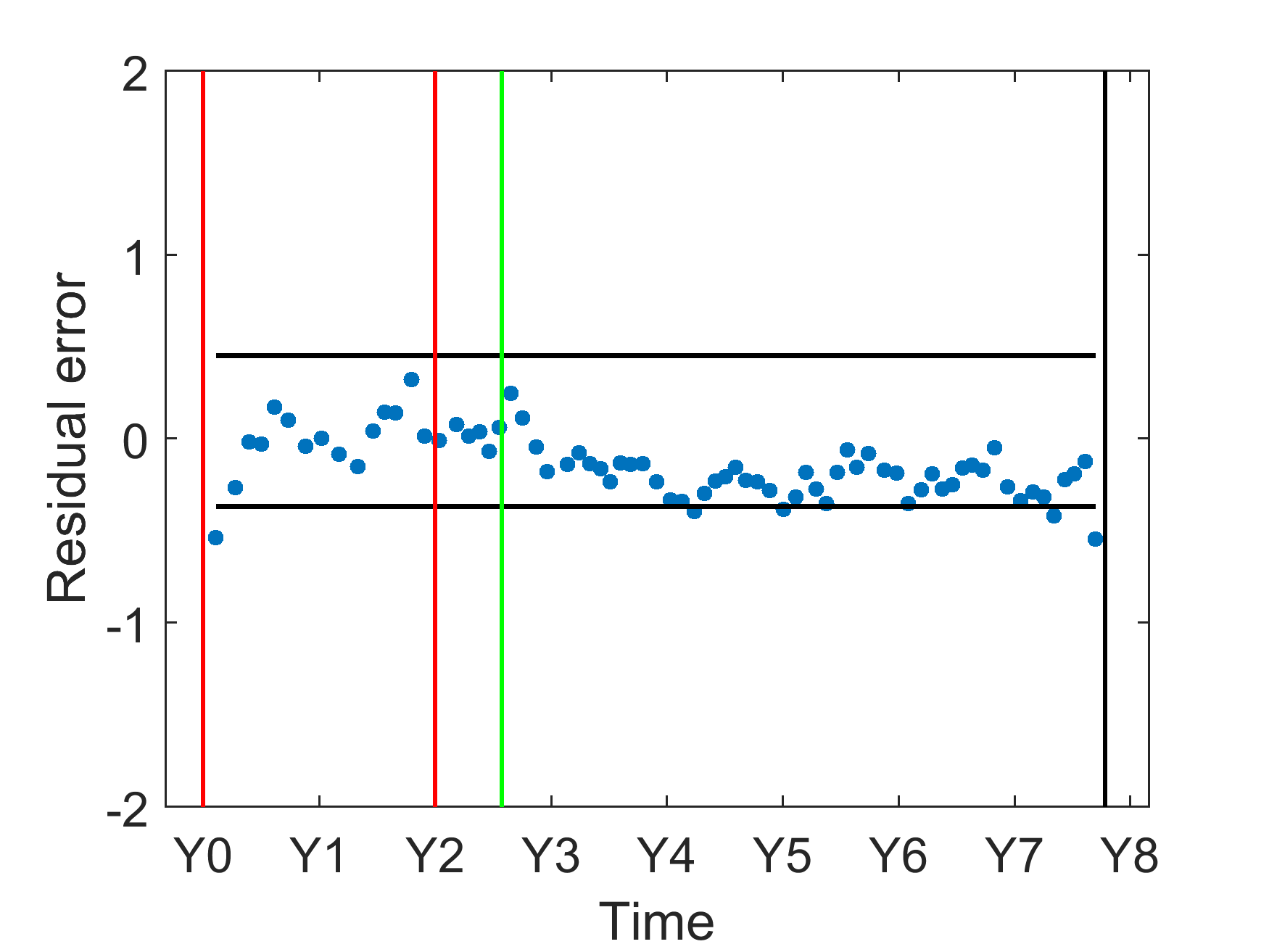}}
    \hspace{0in}
  \vspace{0in}
  \caption{X-bar control charts of GP residuals for: (a) Blade A, (b) Blade B, and (c) Blade C in Site B.}
  \label{fig:x_bar_siteB}
\end{figure}

Figure \ref{fig:x_bar_siteB} shows the results from this final case study. Once again, the 3$\sigma$ thresholds were exceeded in advance, this time only 3 months before the remedial activities to fix the blade took place. Furthermore, the amplitude of the threshold exceedance is not as pronounced as that seen in Example 2. This is primarily attributed to the noisy features used.

\section{Conclusions}

In this article, a new diagnostic methodology for operational wind turbine blades has been proposed. Gaussian Processes (GPs) are used to predict the edge frequencies of one blade, given the edge frequencies of another blade and the ambient temperature. This may be viewed as another way of applying the Johansen procedure to cointegrate variables, as shown in \cite{cross2011cointegration}. The premise is that as long as no damage has taken place on the blade, the relationship between the edge frequencies of the respective blades should remain consistent. However, when damage does take place, the predictive capabilities of the GPs would be affected, and the predictions would not be accurate. Hence, when damage takes place, the GP residuals (the difference between the actual and predicted edge frequencies) would grow in amplitude, and would hence act as informative damage indicators. X-bar control charts, along with 3$\sigma$ thresholds, were used to indicate whether the blades were healthy or not.

A case study using synthesised data was used to illustrate the algorithm. Two other case studies from real turbines, where one of the blades was reported to sustain damage, were also shown to illustrate the diagnostic capabilities of the methodology proposed in this article. The results showed the successful implementation of this method whereby the respective damages were detected in advance of critical failure.

The key outcomes of this work are:

\begin{itemize}
\item Although the individual edge frequency time series of the blades are nonstationary, there exists a common and explicable trend between them, which can be learned.
\item Due to the noisy estimation of the edge frequencies, there may not be a one-to-one mapping between the edge-frequency estimates of the pairs of blades. However, as long as the nominal trend of this correlation is captured by the GPs, the residual errors give sufficient information regarding the onset of damage as long as these frequencies being monitored are sensitive to the damage type and location.
\item The addition of ambient temperature as a feature helps in the predictive capabilities of the GPs, especially when the edge-frequency estimates are noisy. 
\item The methodology of identifying correlations between two (or more) blades may be applied to different types of signals (for example, strains and displacements), although such signals may require additional pre-processing and other complementary features.
\end{itemize}

\section*{Acknowledgements}

The authors would like to thank Siemens Gamesa Renewable Energy for funding this research; without their input, this project would not have been possible. The authors would also like to acknowledge the support of the following EPSRC grants: EP/R004900/1, EP/S001565/1, and EP/R003645/1.

\bibliography{references}

\end{document}